\pdfoutput=1

\documentclass[11pt]{article}

\usepackage[]{acl}

\usepackage{times}
\usepackage{latexsym}

\usepackage[T1]{fontenc}

\usepackage[utf8]{inputenc}

\usepackage{microtype}

\usepackage{enumitem}
\usepackage{amsmath}
\usepackage{booktabs}
\usepackage{graphicx}
\usepackage{hyperref}
\usepackage{multirow}
\usepackage{arydshln}
\usepackage{caption}
\usepackage{subcaption}
\usepackage{booktabs}
\usepackage{longtable}

\title{ERNIE-Code: Beyond English-Centric Cross-lingual Pretraining for Programming Languages}

 \author{Yekun Chai \, Shuohuan Wang \, Chao Pang\, Yu Sun \, Hao Tian \, Hua Wu \\
        Baidu \\ \texttt{\{chaiyekun,wangshuohuan,sunyu02\}@baidu.com}
} 

\begin{document}
\maketitle
\begin{abstract}
Software engineers working with the same programming language (PL) may speak different natural languages (NLs) and vice versa, erecting huge barriers to communication and working efficiency. Recent studies have demonstrated the effectiveness of generative pre-training in computer programs, yet they are always English-centric. In this work, we step towards bridging the gap between multilingual NLs and multilingual PLs for large language models (LLMs). We release ERNIE-Code, a unified pre-trained language model for 116 NLs and 6 PLs. We employ two methods for universal cross-lingual pre-training: span-corruption language modeling that learns patterns from monolingual NL or PL; and pivot-based translation language modeling that relies on parallel data of many NLs and PLs. Extensive results show that ERNIE-Code outperforms previous multilingual LLMs for PL or NL across a wide range of end tasks of code intelligence, including multilingual code-to-text, text-to-code, code-to-code, and text-to-text generation. We further show its advantage of zero-shot prompting on multilingual code summarization and text-to-text translation. We release our code and pre-trained checkpoints\footnote{\url{https://github.com/PaddlePaddle/PaddleNLP/tree/develop/model_zoo/ernie-code}}.

\end{abstract}

\section{Introduction}
\label{sec:intro}
Recent trends in generative pre-training of programming languages~\citep{DBLP:conf/emnlp/FengGTDFGS0LJZ20,Chen2021EvaluatingLL,Li2022CompetitionLevelCG} have led to a proliferation of improvements in code intelligence scenarios, including program understanding and generation~\citep{DBLP:conf/emnlp/0034WJH21,DBLP:conf/naacl/AhmadCRC21}. In this context, a transformer-based large language model (LLM) is pre-trained on a large corpus of open source code (\emph{e.g.}, from GitHub) and then finetuned or zero-shotly evaluated on downstream tasks, such as program synthesis~\citep{Austin2021ProgramSW,Fried2022InCoderAG,Nijkamp2022ACP}, code search~\citep{Husain2019CodeSearchNetCE,Li2021CodeRetrieverUA}, clone detection~\citep{DBLP:conf/nips/LuGRHSBCDJTLZSZ21}, and text-to-code generation~\citep{DBLP:conf/emnlp/ClementDTSS20}.

\begin{figure}[]
\begin{center}
\includegraphics[width=\columnwidth]{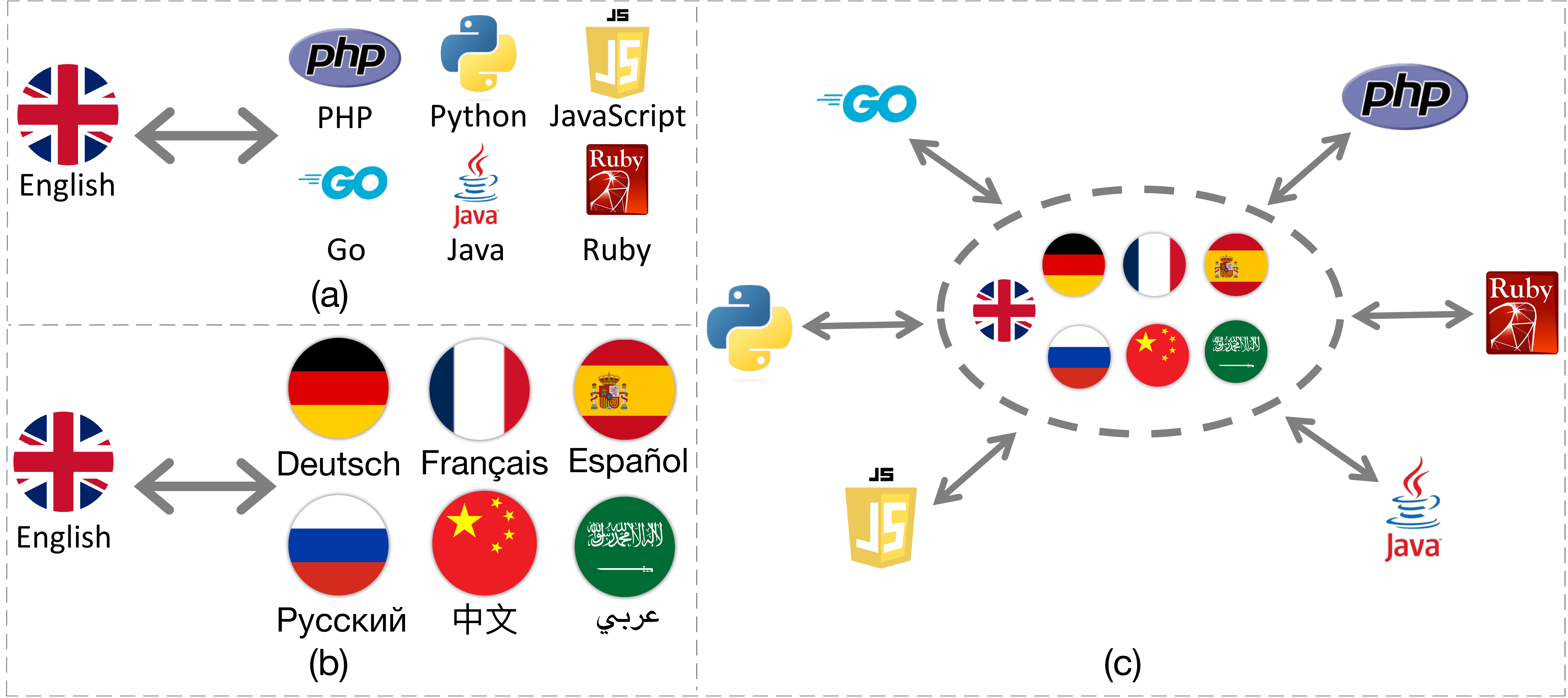}
\caption{Comparison among (a) Multilingual code pre-training; (b) Multilingual text pre-training; (c) Universal multilingual text-code pre-training (ours).}
\label{fig:codem}
\end{center}
\end{figure}

Although there has been a surge of interest in learning general-purpose multilingual LLMs for source code~\citep{DBLP:conf/emnlp/FengGTDFGS0LJZ20,DBLP:conf/naacl/AhmadCRC21,DBLP:conf/emnlp/0034WJH21,Fried2022InCoderAG,Xu2022ASE}, research in this area has been essentially connecting English texts (\emph{e.g.}, comments or docstring) and multiple computer programs (\emph{e.g.}, Python, C++, and Java), as shown in Figure~\ref{fig:codem}(a), and primarily focused around English-centric corpora and benchmarks. This \emph{English-centricity} issue dramatically limits their use and practice given that 95\% of the world population does \emph{not} have English as their native language~\citep{Guo2018NonNativeES}.


As such, it is crucial to mitigate barriers and draw connections between non-English natural languages (NLs) and multiple programming languages (PLs). One engineering solution is to use English translation of non-English texts by engaging neural machine translation (NMT) systems before/after the code LLM as a pipeline. Unfortunately, most general-purpose NMT systems~\citep{Wu2016GooglesNM, Johnson2017GooglesMN} are not designed for code-specific texts and can be prone to accumulative errors due to cascaded prediction stages. 

A more general way is to learn a multilingual LLM that encodes a mixture of multiple NLs and PLs into a shared cross-mode representation space. The success in learning universal representations of many languages~\citep{DBLP:conf/nips/ConneauL19, DBLP:conf/naacl/XueCRKASBR21, DBLP:conf/naacl/AhmadCRC21,DBLP:conf/emnlp/0034WJH21,Xu2022ASE} that focuses on PLs or NLs suggests that it is possible to build a universal multilingual model that jointly represent multiple PLs and NLs. 


In this work, we present ERNIE-Code, a unified cross-lingual pre-trained LLM for multiple NLs and PLs in hopes of mitigating the \textit{English-centric} bias for program pre-training, as illustrated in Figure~\ref{fig:codem}. Our model builds on the T5~\cite{RaffelSRLNMZLL20} encoder-decoder architecture that has been demonstrated to be effective in understanding and generation tasks for multilingual NL~\cite{DBLP:conf/naacl/XueCRKASBR21} and PL~\cite{DBLP:conf/emnlp/0034WJH21}. For monolingual pre-training on mono-mode data (\emph{i.e.}, unpaired multilingual code or text), we follow the same T5 recipe to employ the ``span-corruption'' denoising objective in the text-to-text format. 

The good-quality parallel corpus between low-resource NLs and multilingual PLs is usually unavailable. Instead, most popular PLs, accompanying API documentation, code examples, and discussion forums are primarily written in English, which poses a bottleneck to drawing connections between low-resource NLs and PLs. Inspired by the pivot-based machine translation~\citep{Gispert2006CatalanEnglishSM,utiyama-isahara-2007-comparison} that uses a \textit{pivot} language and decomposes the source$\leftrightarrow$target translation into source$\leftrightarrow$pivot and pivot$\leftrightarrow$target bilingual translation, we introduce the pivot-based translation language modeling (PTLM) with prompting that disassembles multi-NL$\leftrightarrow$multi-PL into multi-NL$\leftrightarrow$English and English$\leftrightarrow$multi-PL with pivoting through English.

Specifically, we leverage the PTLM training in dual direction for parallel corpus in different modes: (1) English$\leftrightarrow$multi-PL. For multi-PL$\leftrightarrow$English parallel data, \emph{i.e.}, code snippets and their accompanying comments, the model learns to generate English comments from code fragments and vice versa. (2) English$\leftrightarrow$Multi-NL. It learns to translate between English and other NLs. The model thus encodes PL$\leftrightarrow$English and English$\leftrightarrow$NL at the same time, with English as a \emph{pivot} language. 
We conduct extensive experiments on different downstream tasks: (1) Multilingual text-to-code generation; (2) Multilingual code summarization (code-to-text); (3) Documentation translation (text-to-text); (4) Code repair (code-to-code). Empirical results have shown that our model outperforms strong multilingual LLMs for PL or NL and have verified its universal multilingual capacity. We also provide examples to show its decent zero-shot capability on code summarization and text translation via zero-shot prompting.






To summarize, this paper makes the following contributions: (1) We first propose a unified cross-lingual pre-trained LLM for both multilingual NLs and multilingual PLs, enlarging the capacity of LLMs towards jointly learning the universal multilingualism.
(2) We employ the pivot-based translation language modeling with prompting to build connections between multi-NLs and multi-PLs (with English pivots) and mitigate the problem when the parallel corpus of multilingual-NL$\leftrightarrow$multilingual-PL is unavailable.
(3) We obtain superior performance compared with previous multilingual LLMs across a wide range of code intelligence tasks, including text-to-code, code-to-text, code repair, and code documentation translation.
(4) To some extent, our model has shown zero-shot prompting ability on multilingual code-to-text, text-to-code, and text-to-text generation. Moreover, ERNIE-Code is well-performed at naming a function and completing corresponding arguments given multilingual NL instructions.



\section{Related work}
\label{sec:bg}

As text-based formal languages with strict syntax and semantics, PL differs from NL because NL is only used for human communication while PL requires the interaction between humans and computers. This work targets bridging the gap between human languages and computer programs in a cross-lingual manner for unified multilingual pre-training, which is closely related to LLMs in either multilingual PL or NL.

\paragraph{Multilingual PL pre-training} 
The success of large-scale pre-training has led to impressive advances in computer programs. This line of research involves pre-training on multilingual PLs using bidirectional transformer encoders~\citep{DBLP:conf/emnlp/FengGTDFGS0LJZ20, Li2021CodeRetrieverUA}, casual transformer decoders~\citep{Chen2021EvaluatingLL,Austin2021ProgramSW,Fried2022InCoderAG,Nijkamp2022ACP,Xu2022ASE}, and transformer encoder-decoder architectures~\citep{DBLP:conf/emnlp/0034WJH21,DBLP:conf/naacl/AhmadCRC21,Li2022CompetitionLevelCG}. Those with bidirectional encoder focus on program understanding tasks, such as code search~\citep{Husain2019CodeSearchNetCE}, while the encoder-decoder ones target at building unified LLMs for both program understanding and generation. We observe that a large body of pre-trained models for PL tend to scale up their parameters under the framework of causal language modeling, mainly focusing on program synthesis~\citep{Chen2021EvaluatingLL,Austin2021ProgramSW,Fried2022InCoderAG,Nijkamp2022ACP,Xu2022ASE}. Nevertheless, all of these works are almost \textit{English-centric}, posing significant challenges to coping with PL end-tasks in non-English scenarios.

\paragraph{Multilingual NL pre-training} This work is also related to the continual trend of multilingual LLMs. One line of this work focuses on encoding multiple NLs into a shared representation space~\citep{DBLP:conf/nips/ConneauL19,DBLP:conf/acl/ConneauKGCWGGOZ20}, while some make efforts to extend the efficient monolingual pre-training method into multilingual settings~\citep{DBLP:conf/naacl/XueCRKASBR21,liu-etal-2020-multilingual-denoising}.


Inheriting the recent success of LLMs in multilingualism, this work lies in the intersection between multilingual NL and PL pre-training. In contrast to the previous work that attends to either multilingual NL or multilingual PL, we seek to explicitly learn multiple NLs and PLs in a shared representation space in hopes of breaking the language barriers between these two modes.

\section{Cross-lingual NL-PL pre-training}
\label{sec:method}
In this section, we introduce pre-training tasks (\S\ref{sec:objective}), model (\S\ref{sec:model}), and pre-training data (\S\ref{sec:data}) we use  throughout this work. 


\subsection{Pre-training tasks}
\label{sec:objective}
We pre-train on two pre-training tasks using both PL and NL data: one (\S\ref{sec:sclm}) uses monolingual PL/NL data (unsupervised), while the other (\S\ref{sec:ptlm}) requires parallel NL-PL and NL-NL pairs (supervised). The former advances to learn intra-modal patterns from PL or NL only, while the latter endows the model with cross-lingual/modal alignment and zero-shot capabilities.

\subsubsection{Task\#1: Span-corruption language modeling (SCLM)}
\label{sec:sclm}
Denoising sequence-to-sequence pre-training has been highly effective across a broad set of tasks, including natural language processing~\citep{liu-etal-2020-multilingual-denoising,RaffelSRLNMZLL20, DBLP:conf/naacl/XueCRKASBR21} and programming language processing~\citep{DBLP:conf/emnlp/0034WJH21, DBLP:conf/naacl/AhmadCRC21}. 
The denoising pre-training objective first corrupts input sequences by masking or adding noise; and then recovers the original inputs by forcing the model to predict corrupted spans, sentences, or documents. \citet{RaffelSRLNMZLL20} finds that span-corruption denoising pre-training produces strong performance while being more computationally efficient on account of shorter target sequence lengths. 

\begin{figure}[h]
\vskip -2mm
\begin{center}
\includegraphics[width=.8\columnwidth]{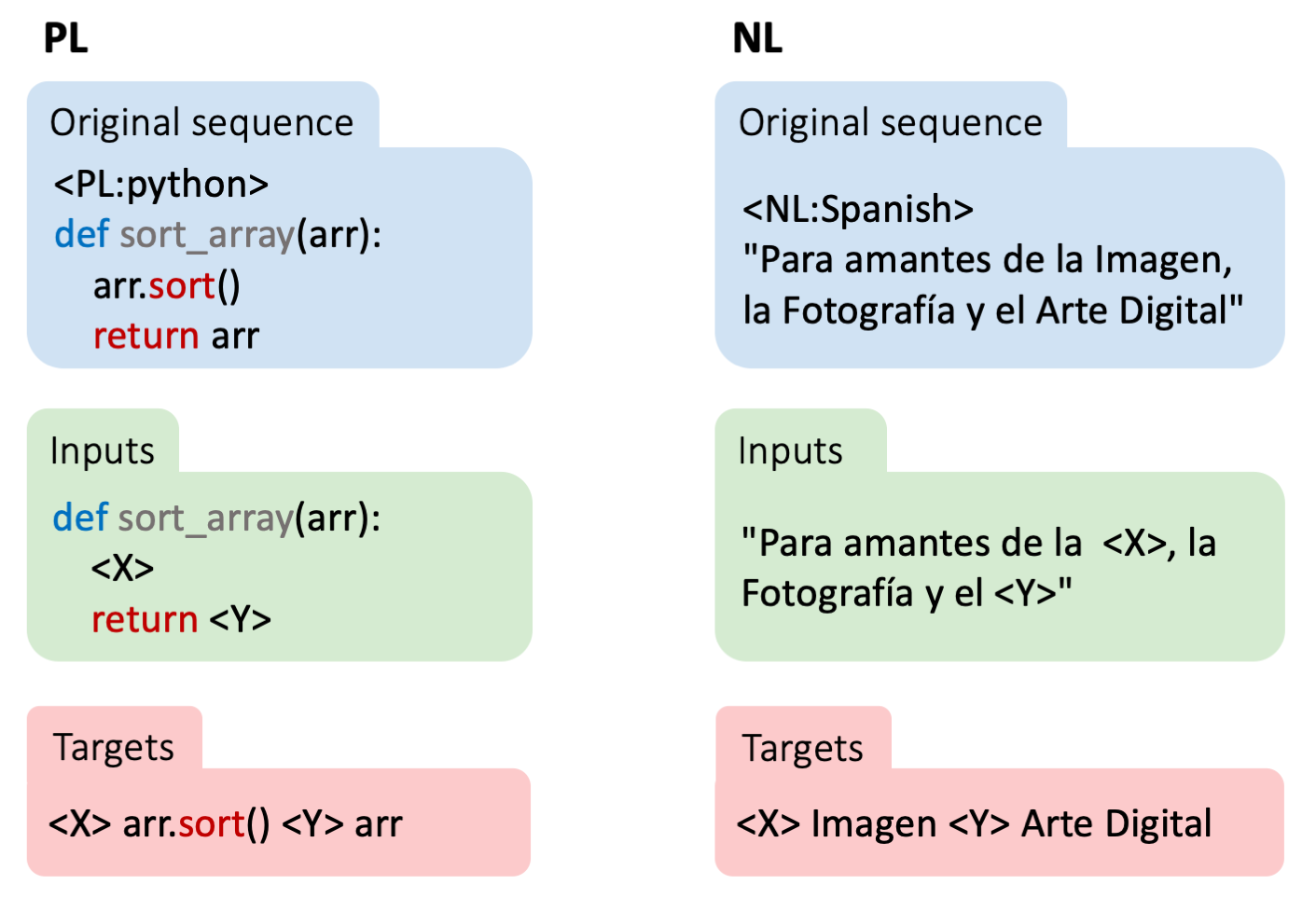}
\vskip -1mm
\caption{Schematic of the SCLM objective for PL (left) and NL (right) example.}
\label{fig:sclm_ex}
\end{center}
\vskip -3mm
\end{figure}

In similar vein, we extend the span-corruption denoising pre-training on both PL and NL. We refer to this task as span-corruption language modeling (SCLM), as illustrated in Figure~\ref{fig:sclm_ex}. Specifically, it corrupts 15\% of the original NL/PL input tokens with a mean span length of 3 by replacing contiguous, randomly-spaced spans of tokens as a single mask placeholder and then predicting the corrupted span on the target side. 

Suppose we have a total of $M$ monolingual corpora of NL and PL corpora $\{ C_m\}_{m=1 \cdots M}$. We apply the SCLM pre-training objective on both NL and PL data in a multi-tasking fashion:
\begin{align}
    \mathcal{L}_\textrm{SCLM} &{}= \sum_{m=1}^M \sum_{t=1}^T - \log P_\theta \big(x_{(i),t} | \mathbf{x}_{(m)}^{\backslash\textrm{mask}}, \mathbf{x}_{(m),<t}^{\textrm{mask}} \big) 
\end{align}{
where $\theta$ denotes trainable parameters, $\mathbf{x}_{(m)}^{\backslash\textrm{mask}}$ and $\mathbf{x}_{(m)}^{\textrm{mask}}$ are span-corrupted inputs 
and corresponding target spans from monolingual corpus $C_m$, respectively. $\mathbf{x}_{(m),<t}^{\textrm{mask}}$ indicates the generated tokens until the $t$-th time step out of the target (corrupted) sequence length $T$.
}

\subsubsection{Task\#2: Pivot-based translation language modeling (PTLM)}
\label{sec:ptlm}
This work aims at narrowing the cross-modal cross-lingual gap between multiple NLs and PLs, yet good quality parallel corpora between non-English NL and multilingual PL are unavailable. The lack of parallel corpus stems from the fact that most popular PLs, accompanying documentations, and discussion websites are primarily written in English. Early investigation of statistical machine translation proposed pivot-based approach~\citep{Gispert2006CatalanEnglishSM, utiyama-isahara-2007-comparison} to introducing a third language - named \emph{pivot} language - for which there exist good-quality source-pivot and pivot-target bilingual corpora. \citet{Johnson2017GooglesMN} adopt a single NMT model to simultaneously learn many translation directions (including source$\leftrightarrow$pivot, pivot$\leftrightarrow$target), enabling the zero-shot translation between NLs implicitly. 

 
In our context, the good-quality multi-PL to the multi-NL bilingual corpus is unavailable, yet there exists multi-NL to English and English to multi-PL parallel corpora, with pivoting through English. Motivated by the pivot-based NMT~\citep{Johnson2017GooglesMN} and translation language modeling (TLM; ~\citealp{DBLP:conf/nips/ConneauL19}) approach, we apply a unified pivot-based training objective to the course of multilingual NL-PL pre-training, namely pivot translation language modeling (PTLM).

\begin{figure}[h]
\vskip -2mm
\begin{center}
\includegraphics[width=\columnwidth]{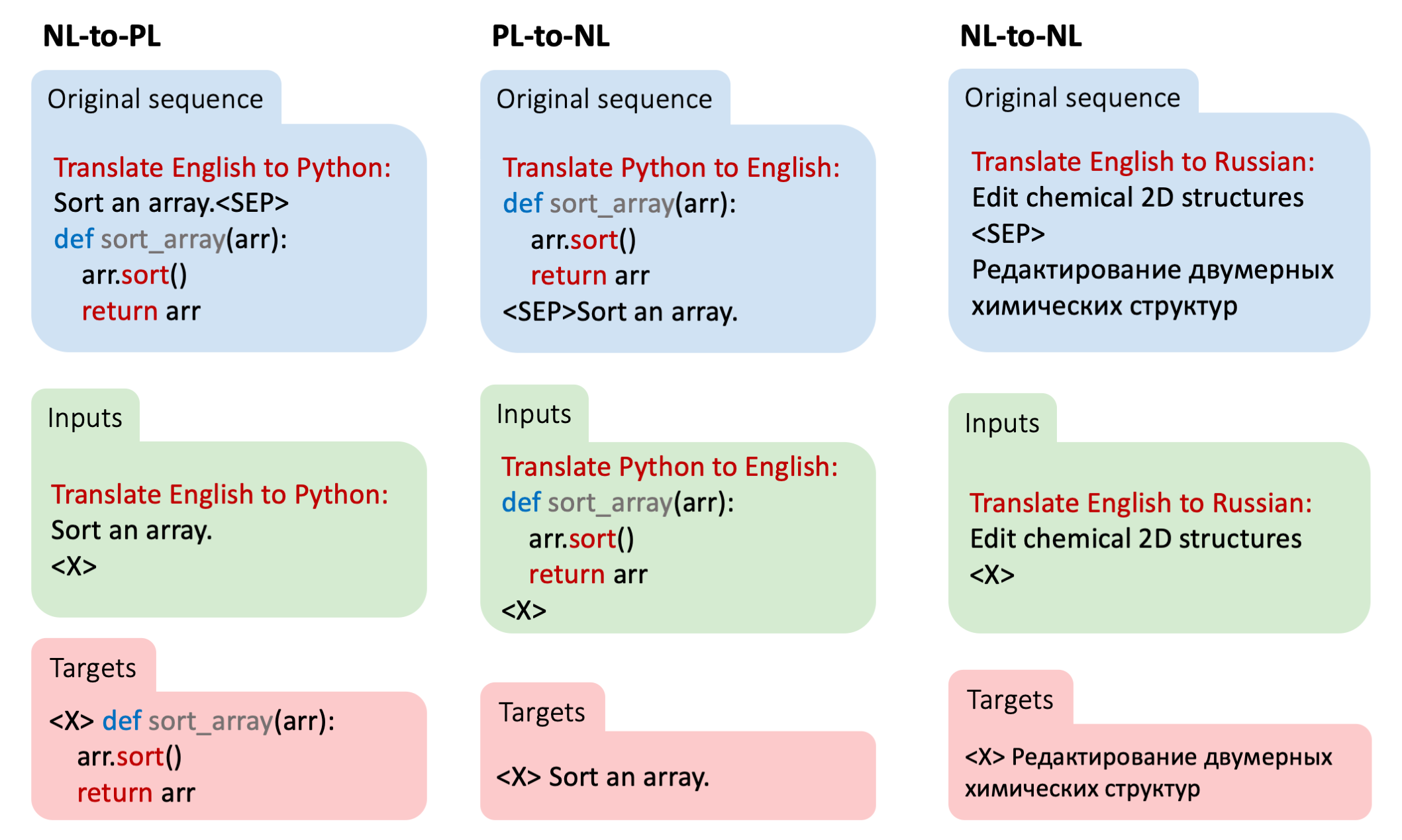}
\vskip -1mm
\caption{Schematic of the PTLM objective for NL-to-PL (left), PL-to-NL (middle), NL-to-NL (right) example. ``<SEP>'' indicates the delimiter token.}
\label{fig:ptlm_ex}
\end{center}
\vskip -3mm
\end{figure}

With bilingual PL-NL and NL-NL corpora, we jointly learn the parallelism with pivoting in dual directions: for instance, Python$\leftrightarrow$English and English$\leftrightarrow$Russian. This allows for implicit bridging between PL-NL pairs that are never seen explicitly in training data~\citep{Johnson2017GooglesMN}. 
More precisely, we concatenate parallel source-target sentences and learn to predict the corrupted target language, as shown in Figure~\ref{fig:ptlm_ex}. Instead of masking random tokens \citep{DBLP:conf/nips/ConneauL19}, we corrupt the \textit{whole} sentence in either direction of bilingual data and predict on the target side. The model requires attending to complete representations of source sentences to recover the target sentence and learn the alignment between source-target pairs. 
Suppose we have $N$ bilingual NL-NL and NL-PL parallel corpora $\{D_n\}_{n=1,\cdots,N}$. We can formulate the PTLM training as:

\begin{align}
    \mathcal{L}_\textrm{PTLM} &{}= \sum_{n=1}^N \sum_{t=1}^T - \log P_\theta \big(x_{(n),t} | \mathbf{x}_{(n)}^{\textrm{source}}, \mathbf{x}_{(n),<t}^{\textrm{target}} \big) 
\end{align}{where $\mathbf{x}_{(n)}^{\textrm{source}}$ and $\mathbf{x}_{(n)}^{\textrm{target}}$ denote source 
and target sentences from bilingual corpus $D_n$. $\mathbf{x}_{(n),<t}^{\textrm{target}}$ indicates the generated tokens until the $t$-th time step out of the target sequence length $T$.
This training format is the same as an NMT task.
}

To enable a pivot-based approach and specify the target language, we reformat the PTLM by prompting with a task prefix (See Figure~\ref{fig:ptlm_ex}), in which we prepend a task instruction ``translate \verb|A| to \verb|B|: \verb|\n|'' on the left of input sentences, where \verb|A| and \verb|B| denote the source and target language, respectively. This prompt instruction indicates the target language the model should translate to, resulting in descent zero-shot abilities (\S\ref{sec:zero-shot}).

\subsection{Model}
\label{sec:model}

\paragraph{Model architecture} 
Our model follows the same architecture as T5-base~\citep{RaffelSRLNMZLL20}. Specifically, we build ERNIE-Code on ``T5.1.1'' version\footnote{\url{https://github.com/google-research/text-to-text-transfer-transformer/blob/main/released_checkpoints.md\#t511}}, which improves upon T5 using gated nonlinearities~\citep{Shazeer2020GLUVI, chai-etal-2020-highway}.  We refer to \S\ref{ap:pretrain} for pre-training settings.




\paragraph{Shared NL/PL encoding} We base our tokenizer on SentencePiece tokenizer in \citet{DBLP:conf/naacl/XueCRKASBR21}. However, the original SentencePiece tokenizer designed for encoding NLs does not effectively represent PL data. We thus add a set of tokens representing whitespace indentation of different lengths in PL. See tokenization details in \S\ref{ap:tokenization}.


\subsection{Pre-training data}
\label{sec:data}

\paragraph{Code corpus} For PL data, we use the same pre-training corpora - CodeSearchNet~\citep{Husain2019CodeSearchNetCE} - as previous models~\citep{DBLP:conf/emnlp/FengGTDFGS0LJZ20,DBLP:conf/emnlp/0034WJH21}.\footnote{Note that for a fair comparison, we do not use additional data from public repositories hosted on GitHub.} It covers six monolingual PLs (Go, Java, JavaScript, PHP, Python, and Ruby) and six NL-PL parallel data, \emph{i.e.}, PL-NL query pairs. The majority of NL annotations in the parallel corpora is English. We defer data statistics and preprocessing details in \S\ref{ap:pl_data}.

\paragraph{Text corpus} We pre-train on the following NL data corpus:  (1) Monolingual data from CC-100~\citep{DBLP:conf/acl/ConneauKGCWGGOZ20} that built on a clean CommonCrawl corpus\footnote{\url{https://data.statmt.org/cc-100}}, containing 116 different NLs. \footnote{Note that following~\citet{DBLP:conf/acl/ConneauKGCWGGOZ20}, we count Romanized variants as separate languages.} (2) Parallel data from OPUS website\footnote{\url{https://opus.nlpl.eu}} covering 15 languages. The collected NL translation pairs include MultiUN~\citep{DBLP:conf/lrec/ZiemskiJP16}, IIT Bombay~\citep{DBLP:conf/lrec/KunchukuttanMB18}, OPUS~\citep{DBLP:conf/lrec/Tiedemann12}, WikiMatrix~\citep{DBLP:conf/eacl/SchwenkCSGG21}, \emph{etc}. We refer to \S\ref{ap:nl_data} for details.

To alleviate the bias towards high-resource languages, we follow~\citet{DBLP:conf/nips/ConneauL19} to rebalance the data distribution on both corpora and up/down-sample sentences from each language (or language pair) $i$ with a rescaled multinomial distribution $q_i$:
\begin{align} \label{eq:rescale}
    q_i &{}= \frac{p_i^\alpha}{\sum_{j=1} p_j^\alpha} 
\end{align}{where $p_i$ is the data percentage of each monolingual or parallel corpus. Following ~\citet{DBLP:conf/nips/ConneauL19}, we set $\alpha=0.3$ for both monolingual and parallel corpus.}

\section{Experiments}
\label{sec:exp}

In this section, we first introduce multilingual pre-trained models for comparison (\S\ref{sec:baselines}), downstream tasks, and evaluation metrics (\S\ref{sec:task}). Then we evaluate and show consistent performance gains on several multilingual NL/PL benchmarks, including code-to-text (\S\ref{sec:c2t}), text-to-code (\S\ref{sec:t2c}), text-to-text (\S\ref{sec:t2t}), and code-to-code (\S\ref{sec:c2c}) end tasks. 



\subsection{Comparison to related models}
\label{sec:baselines}

\begin{table}[]
\vskip -3mm
\centering
\resizebox{\columnwidth}{!}{%
\begin{tabular}{@{}llrrl@{}}
\toprule
Model                                              & \#Param & \#PLs & \#NLs & Data source                \\ \midrule
mBART~\citep{liu-etal-2020-multilingual-denoising} & 680M      & -     & 25    & Common Crawl (CC25)        \\
mT5~\citep{DBLP:conf/naacl/XueCRKASBR21}           & 560M      & -     & 101   & Common Crawl (mC4)         \\
PLBART~\citep{DBLP:conf/naacl/AhmadCRC21}          & 390M      & 2     & 1     & GitHub,  StackOverflow     \\
CodeT5~\citep{DBLP:conf/emnlp/0034WJH21}           & 220M      & 8     & 1     & CodeSearchNet, GitHub (C/C\#) \\ \hdashline
ERNIE-Code (ours)                                       & 560M      & 6     & 116   & CodeSearchNet, CC-100, OPUS \\ \bottomrule
\end{tabular}%
}
\vskip -1mm
\caption{Comparison of our model to existing massively multilingual pre-trained models for NLs and PLs.}
\label{tab:baselines}
\vskip -4mm
\end{table}

To contextualize our new model, we briefly compare it with existing multilingual LLMs for NLs/PLs. Considering that ERNIE-Code is the first LLM targeting multilingual NL and PL explicitly, for brevity, we focus on models that support either many NLs or many PLs. Table~\ref{tab:baselines} reports the overall statistics of comparison models.

\textbf{mBART}~\citep{liu-etal-2020-multilingual-denoising} is a multilingual-NL variant of BART~\citep{lewis-etal-2020-bart} trained with a full-text denoising objective on a subset of 25 languages from CommonCrawl. It learns to reconstruct the full NL texts from corrupted ones with an arbitrary noising function. \textbf{mT5}~\citep{DBLP:conf/naacl/XueCRKASBR21} is a multilingual-NL encoder-decoder model adapted from T5. It is trained on 101 NLs using filtered CommonCrawl data (mC4) using the same SCLM objective as our model. \textbf{PLBART}~\citep{DBLP:conf/naacl/AhmadCRC21} is a multilingual-PL version of BART with a denoising objective using three noising formats. It is trained on 210M Java functions, 470M Python functions from GitHub, and 47M English posts from StackOverflow. \textbf{CodeT5}~\citep{DBLP:conf/emnlp/0034WJH21} is a PL version of mT5 that is pre-trained on six-PL monolingual/parallel data from CodeSearchNet and extra C/C\# data collected from GitHub. It additionally learns token-type information from identifiers and applies dual generation between English and PLs.

\begin{table*}[]
\centering
\vskip -3mm
\resizebox{\textwidth}{!}{%
\begin{tabular}{crrrrrrrrrrrr}
\hline\hline
\multicolumn{1}{c|}{\multirow{2}{*}{\textbf{Model}}} & \multicolumn{3}{c|}{\textbf{Spanish}} & \multicolumn{3}{c|}{\textbf{Japanese}} & \multicolumn{3}{c|}{\textbf{Russian}} & \multicolumn{3}{c}{\textbf{Avg.}} \\ \cline{2-13} 
\multicolumn{1}{c|}{} & \multicolumn{1}{l}{\textbf{B-4}} & \multicolumn{1}{l}{\textbf{R-L}} & \multicolumn{1}{l|}{\textbf{chrF}} & \multicolumn{1}{l}{\textbf{B-4}} & \multicolumn{1}{l}{\textbf{R-L}} & \multicolumn{1}{l|}{\textbf{chrF}} & \multicolumn{1}{l}{\textbf{B-4}} & \multicolumn{1}{l}{\textbf{R-L}} & \multicolumn{1}{l|}{\textbf{\textbf{chrF}}} & \multicolumn{1}{l}{\textbf{B-4}} & \multicolumn{1}{l}{\textbf{R-L}} & \multicolumn{1}{l}{\textbf{chrF}} \\ \hline
\multicolumn{13}{c}{\textbf{Translate-train}} \\ \hline
\multicolumn{1}{c|}{\textbf{mBART}} & 0.96 & 19.46 & \multicolumn{1}{r|}{19.30} & 0.07 & 4.70 & \multicolumn{1}{r|}{7.88} & 0.08 & 0.00 & \multicolumn{1}{r|}{13.56} & 0.37 & 8.05 & 13.58 \\
\multicolumn{1}{c|}{\textbf{mT5}} & 0.94 & 28.69 & \multicolumn{1}{r|}{19.87} & 0.06 & 2.95 & \multicolumn{1}{r|}{6.58} & 0.09 & 2.56 & \multicolumn{1}{r|}{12.00} & 0.36 & 11.40 & 12.82 \\
\multicolumn{1}{c|}{\textbf{PLBART}} & 0.16 & 14.33 & \multicolumn{1}{r|}{11.72} & 0.06 & 4.11 & \multicolumn{1}{r|}{7.87} & 0.24 & 2.98 & \multicolumn{1}{r|}{14.06} & 0.15 & 7.14 & 11.22 \\
\multicolumn{1}{c|}{\textbf{CodeT5}} & 1.00 & 22.93 & \multicolumn{1}{r|}{20.09} & 0.04 & 5.42 & \multicolumn{1}{r|}{7.13} & 0.13 & 1.48 & \multicolumn{1}{r|}{12.97} & 0.39 & 9.94 & 13.40 \\ \hdashline
\multicolumn{1}{c|}{\textbf{Ours(L512)}} & 1.90 & 32.51 & \multicolumn{1}{r|}{23.22} & 0.30 & \textbf{10.62} & \multicolumn{1}{r|}{\textbf{9.16}} & \textbf{0.43} & 5.01 & \multicolumn{1}{r|}{\textbf{16.60}} & 0.88 & \textbf{16.05} & \textbf{16.33} \\
\multicolumn{1}{c|}{\textbf{Ours(L1024)}} & \textbf{2.51} & \textbf{33.87} & \multicolumn{1}{r|}{\textbf{24.00}} & \textbf{0.58} & 8.55 & \multicolumn{1}{r|}{8.81} & 0.28 & \textbf{5.69} & \multicolumn{1}{r|}{15.24} & \textbf{1.12} & 16.04 & 16.02 \\ \hline
\multicolumn{13}{c}{\textbf{Zero-shot}} \\ \hline
\multicolumn{1}{c|}{\textbf{Ours(L512)}} & 0.49 & 12.78 & \multicolumn{1}{r|}{15.69} & \textbf{1.46} & \textbf{32.07} & \multicolumn{1}{r|}{\textbf{11.02}} & \textbf{1.98} & \textbf{30.46} & \multicolumn{1}{r|}{11.68} & \textbf{1.31} & \textbf{25.10} & 12.80 \\ \hline\hline
\end{tabular}%
}
\vskip -2mm
\caption{Results of multilingual code summarization task. ``L512/1024'' indicates the maximum length of 512/1024.}
\label{tab:c2t}
\vskip -3mm
\end{table*}

\subsection{Evaluation datasets and metrics}
\label{sec:task}
Table~\ref{tab:datasets} displays the statistics of evaluation dataset. We use the same public datasets and train-test splits for all downstream tasks. We refer to \S\ref{ap:finetune} for experimental settings of finetuning.

\textbf{Multilingual code summarization} is a code-to-text task that aims to generate multilingual texts given a code snippet. We use mCoNaLa~\citep{Wang2022MCoNaLaAB} to evaluate the performance of generating multilingual NL from PL. It consists of 341/210/345 manually curated parallel samples with NL in Spanish/Japanese/Russian and PL in Python. As mCoNaLa does not provide the training and validation set, we use CoNaLa~\citep{yin2018mining}, an English-Python parallel data (consisting of \#2,379 samples), as the train/dev set (with 10:1 data split) after translation. For ``\textit{translate-train}'' settings, we use machine-translated CoNaLa as training and dev sets, while use mCoNaLa as the test set. Particularly, we translate CoNaLa's training set into three target languages using FLORES-101~\citep{goyal-etal-2022-flores} and adopt them as train/dev set. We utilize ROUGE-L (R-L; \citealp{lin-2004-rouge}), BLEU-4 (B-4; \citealp{post-2018-call}), and chrF~\citep{popovic-2015-chrf} for comprehensive comparison. 


%

\textbf{Multilingual text-to-code generation} refers to the code generation task that generates code fragments from multilingual NL instructions. We use the same train/dev/test set as the code summarization mentioned above. Specifically, under ``\textit{translate-train}'' settings, we use translated CoNaLa data as training and dev set, mCoNaLa as the test set to generate Python code from NL instruction in three different NLs (\emph{i.e.}, Spanish, Japanese, and Russian). We use ROUGE-L, BLEU-4, and CodeBLEU (C-B;~\citealp{Ren2020CodeBLEUAM}) for evaluating code predictions.



\textbf{Documentation translation} is a text-to-text task that translates code documentation from one NL to another. We use Microsoft Docs from CodeXGLUE dataset~\citep{Lu2021CodeXGLUEAM} to verify the multilingual NL translation between English $\leftrightarrow$ Danish, Latvian, Norwegian, and Chinese. We report BLEU-4 and exact match (EM) in our results.

\textbf{Code repair} is a code-to-code task that automatically fixes bugs given a piece of buggy code. We evaluate on Bugs2Fix \citep{Tufano2019AnES} dataset with two subsets: (i) ``small'' with tokens less than 50; (ii) ``medium'' with a length of between 50 and 100. We report BLEU-4\footnote{\url{https://github.com/microsoft/CodeXGLUE/blob/main/Code-Code/code-refinement/evaluator/evaluator.py}} and EM for evaluation.

\subsection{Multilingual code summarization}
\label{sec:c2t}
Table~\ref{tab:c2t} shows the multilingual code-to-text results of generated NL summaries in Spanish, Japanese, and Russian. We use translated English CoNaLa as training sets in target three languages\footnote{\url{https://conala-corpus.github.io/}}, denoted as ``translate-train'' evaluation.
As shown in Table~\ref{tab:c2t}, our model outperforms all baseline LLMs for either NL (mBART, mT5) or PL (PLBART, CodeT5). In particular, ERNIE-Code, with a length of 1024, exceeds its counterpart of 512-length (1.12 vs. 0.88 on BLEU-4) in that it allows for learning more extended contexts from training NL/PL segments. PLBART performs worst among all baselines on average, while CodeT5, mT5, and mBART behave similarly. We conjecture that PLBART only learns data from Java/Python functions and English StackOverflow posts, whose training data lacks the diversity of multilingualism.

\begin{table*}[]
\vskip -3mm
\centering
\resizebox{\textwidth}{!}{%
\begin{tabular}{c|rrr|rrr|rrr|rrr}
\hline\hline
\multirow{2}{*}{\textbf{Model}} & \multicolumn{3}{c|}{\textbf{Spanish}} & \multicolumn{3}{c|}{\textbf{Japanese}} & \multicolumn{3}{c|}{\textbf{Russian}} & \multicolumn{3}{c}{\textbf{Avg.}} \\ \cline{2-13} 
 & \multicolumn{1}{l}{\textbf{B-4}} & \multicolumn{1}{c}{\textbf{R-L}} & \multicolumn{1}{l|}{\textbf{C-B}} & \multicolumn{1}{l}{\textbf{B-4}} & \multicolumn{1}{c}{\textbf{R-L}} & \multicolumn{1}{l|}{\textbf{C-B}} & \multicolumn{1}{l}{\textbf{B-4}} & \multicolumn{1}{c}{\textbf{R-L}} & \multicolumn{1}{l|}{\textbf{C-B}} & \multicolumn{1}{l}{\textbf{B-4}} & \multicolumn{1}{c}{\textbf{R-L}} & \multicolumn{1}{l}{\textbf{C-B}} \\ \hline
\multicolumn{13}{c}{\textbf{Translate-train}} \\ \hline
\textbf{mBART} & 1.73 & 11.85 & 0.05 & 3.68 & 10.33 & 0.08 & 2.34 & 9.23 & 0.07 & 2.58 & 10.47 & 0.07 \\
\textbf{mT5} & 0.27 & 3.51 & 0.05 & 0.22 & 2.91 & 0.07 & 0.25 & 6.17 & 0.04 & 0.25 & 4.20 & 0.05 \\
\textbf{PLBART} & 2.19 & 14.47 & \textbf{0.06} & 6.56 & 18.26 & 0.09 & 3.27 & 19.92 & \textbf{0.09} & 4.01 & 17.55 & \textbf{0.08} \\
\textbf{CodeT5} & 1.97 & 14.47 & 0.05 & 7.46 & 18.58 & 0.09 & 4.26 & 17.96 & 0.07 & 4.56 & 17.00 & 0.07 \\ \hdashline
\textbf{Ours(L512)} & 2.25 & \textbf{14.92} & \textbf{0.06} & \textbf{8.06} & \textbf{22.65} & \textbf{0.10} & 6.12 & \textbf{25.27} & 0.08 & 5.48 & \textbf{20.95} & \textbf{0.08} \\
\textbf{Ours(L1024)} & \textbf{2.51} & 12.65 & \textbf{0.06} & \textbf{8.08} & 20.12 & 0.09 & \textbf{6.55} & 23.84 & \textbf{0.09} & \textbf{5.71} & 18.87 & \textbf{0.08} \\ \hline
\multicolumn{13}{c}{\textbf{Zero-shot}} \\ \hline
\textbf{Ours(L512)} & 2.47 & 12.12 & \textbf{0.10} &  2.56 & 14.46 & \textbf{0.15} & 3.69 & 13.52  & \textbf{0.14} & 2.91  & 13.37  & \textbf{0.13} \\
\hline\hline
\end{tabular}%
}
\vskip -1mm
\caption{Results of on multilingual text-to-code generation task. }
\label{tab:t2c}
\end{table*}

\subsection{Multilingual text-to-code generation}
\label{sec:t2c}
Table~\ref{tab:t2c} shows the ``translate-train'' results of multilingual text-to-code generation on mCoNaLa. ERNIE-Code outperforms all baselines on BLEU-4, ROUGE-L, and CodeBLEU scores, showing that our multilingual PL-NL pre-training can capture code syntax and semantics. Among all code generation tasks, multilingual models for NL behave worse than those counterparts of PL. PLBART beats all baselines on surface-form n-gram match (BLEU-4/ROUGE-L) and structured code-related match (CodeBLEU), even achieving on par with our model on CodeBLEU. In contrast, mT5 underperforms all the other models on either of three subtasks, suggesting that the mT5 tokenizer is ineffective in encoding PLs, as aforementioned in \S\ref{sec:model}. By comparing mT5 and our models, the improvements suggest our approach's effectiveness in encoding whitespace characters for tokenization. Our model with more extended contexts (1024-length) overshadows that of 512-length on all three text-to-code subtasks.

\begin{table*}[]
\centering
\resizebox{\textwidth}{!}{%
\begin{tabular}{c|cc|cc|cc|cc|cc|cc|cc|cc|cc}
\hline\hline
\multirow{3}{*}{\textbf{Model}} & \multicolumn{4}{c|}{\textbf{En-Da}} & \multicolumn{4}{c|}{\textbf{En-Lv}} & \multicolumn{4}{c|}{\textbf{En-No}} & \multicolumn{4}{c|}{\textbf{En-Zh}} & \multirow{3}{*}{\textbf{\begin{tabular}[c]{@{}c@{}}Avg. \\ B-4\end{tabular}}} & \multirow{3}{*}{\textbf{\begin{tabular}[c]{@{}c@{}}Avg.\\ EM\end{tabular}}} \\ \cline{2-17}
 & \multicolumn{2}{c}{$\rightarrow$} & \multicolumn{2}{c|}{$\leftarrow$} & \multicolumn{2}{c}{$\rightarrow$} & \multicolumn{2}{c|}{$\leftarrow$} & \multicolumn{2}{c}{$\rightarrow$} & \multicolumn{2}{c|}{$\leftarrow$} & \multicolumn{2}{c}{$\rightarrow$} & \multicolumn{2}{c|}{$\leftarrow$} &  &  \\ \cline{2-17}
 & \textbf{B-4} & \textbf{EM} & \textbf{B-4} & \textbf{EM} & \textbf{B-4} & \textbf{EM} & \textbf{B-4} & \textbf{EM} & \textbf{B-4} & \textbf{EM} & \textbf{B-4} & \textbf{EM} & \textbf{B-4} & \textbf{EM} & \textbf{B-4} & \textbf{EM} &  &  \\ \hline
\textbf{Transformer} & 53.31 & - & 58.73 & - & 37.85 & - & 50.37 & - & 53.84 & - & 57.73 & - & 59.90 & - & 50.00 & - & 52.67 & - \\
\textbf{XLM-R} & 67.09 & - & 67.02 & - & 51.92 & - & 68.30 & - & 68.00 & - & 71.84 & - & 70.60 & - & 64.47 & - & 66.16 & - \\
\textbf{mT5} & 67.39 & 10.6 & 68.72 & 24.1 & 57.69 & 8.5 & 64.95 & 22.2 & 68.40 & 12.3 & 68.02 & 23.3 & 72.26 & 20.0 & 68.64 & 24.7 & 67.01 & 18.21 \\
\hdashline
\textbf{Ours(L512)} & \textbf{71.16} & 13.2 & \textbf{72.70} & 27.2 & 60.98 & \textbf{10.6} & 69.28 & 24.3 & \textbf{71.39} & \textbf{15.7} & 72.28 & 26.3 & \textbf{74.53} & 24.3 & 72.43 & \textbf{28.5} & 70.59 & 21.26 \\
\textbf{Ours(L1024)} & 70.90 & \textbf{13.6} & 72.55 & \textbf{27.3} & \textbf{61.30} & \textbf{10.6} & \textbf{69.85} & \textbf{25.1} & 71.11 & \textbf{15.7} & \textbf{72.49} & \textbf{26.7} & 74.49 & \textbf{24.7} & \textbf{72.49} & 28.3 & \textbf{70.65} & \textbf{21.50} \\ \hline\hline
\end{tabular}%
}
\vskip -2mm
\caption{Results of documentation translation. We report BLEU-4 (B-4) and exact match (EM) scores.}
\label{tab:t2t}
\vskip -1mm
\end{table*}


\subsection{Documentation translation (text-to-text)}
\label{sec:t2t}
We further investigate the multilingual text-to-text translation between English (en) and Danish (da)/Latvian (lv)/Norwegian(no)/Chinese(zh). Table~\ref{tab:t2t} shows the documentation translation results of comparison models, including multilingual transformer~\citep{Johnson2017GooglesMN}, XLM-R~\citep{DBLP:conf/acl/ConneauKGCWGGOZ20}, and mT5. Specifically, we finetune our model in a multilingual manner where all bilingual language pairs are learned simultaneously. 

Our model surpasses mT5 and XLM-R in all eight translation directions, demonstrating that our model can perform code-related text-to-text translation. 
As the experiment design only aims to verify the NL translation ability of our model, we did not conduct comprehensive results to compare with state-of-art (SOTA) NMT methods.

\begin{table}[]
\centering
\vskip -2mm
\resizebox{\columnwidth}{!}{%
\begin{tabular}{l|rr|rr}
\hline\hline
\multirow{2}{*}{\textbf{Model}} & \multicolumn{2}{c|}{\textbf{Refine small}} & \multicolumn{2}{c}{\textbf{Refine medium}} \\ \cline{2-5} 
                                & \textbf{B-4}         & \textbf{EM}        & \textbf{B-4}         & \textbf{EM}        \\ \hline
\textbf{Naive copy}             & 78.06                 & 0                  & 90.91                 & 0                  \\
\textbf{RoBERTa (code)}         & 77.30                 & 15.90              & 90.07                 & 4.10               \\
\textbf{CodeBERT}               & 77.42                 & 16.40              & 91.07                 & 5.20               \\
\textbf{PLBART}                 & 77.02                 & 19.21              & 88.50                 & 8.98               \\
\textbf{CodeT5}                 & 78.06                 & \textbf{22.59}              & 88.90                 & \textbf{14.18}              \\ \hdashline
\textbf{Ours (L512)}           & 80.09        & 13.21              & \textbf{91.20}        & 2.22               \\
\textbf{Ours (L1024)}          & \textbf{80.10}        & 12.43              & 91.17                 & 2.00               \\ \hline\hline
\end{tabular}%
}
\vskip -2mm
\caption{Results of program repair task.}
\label{tab:c2c}
\vskip -3mm
\end{table}

\subsection{Program repair (code-to-code)}
\label{sec:c2c}
We further validate that our model can perform code-to-code generation. Table~\ref{tab:c2c} demonstrates the comparison model results on the Bugs2Fix benchmark. Baseline models include RoBERTa (code) - a PL variant of RoBERTa~\citep{Liu2019RoBERTaAR}, CodeBERT~\citep{DBLP:conf/emnlp/FengGTDFGS0LJZ20}, PLBART, and CodeT5. 

On ``small'' and ``medium'' tasks, our model achieves 80.10 and 91.20 BLEU scores, outperforming or achieving competitive results compared with previous SOTA performance.\footnote{Note that EM only serves as a reference indicator in that it is too strict and inaccurate for evaluation, especially for PL hypotheses with the same semantic logic but in various surface forms.} The results of 1024-length and 512-length models slightly differ, possibly because both ``small'' and ``medium'' Java data are of no more than 100-token length, far shorter than our model's length limit.

\section{Analysis}
\label{sec:ana}


\subsection{Syntactic \& semantic probing}
\label{sec:code_probe}
Code fragments with highly-overlapping surface forms but with different semantic and syntactic logic can be given high scores by NL evaluation metrics, such as BLEU and ROUGE. To evaluate the semantic and syntactic aspects of text-to-code generation, we follow~\citet{Ren2020CodeBLEUAM} to adopt dataflow and abstract syntax tree (AST) match to compute the accuracy of dataflow graph and AST subtrees between hypothesis and reference. We refer to~\citet{Ren2020CodeBLEUAM} for further details. 

\begin{figure*}[]
\vskip -2mm
     \centering
     \begin{subfigure}[b]{0.46\textwidth}
         \centering
         \includegraphics[width=\textwidth]{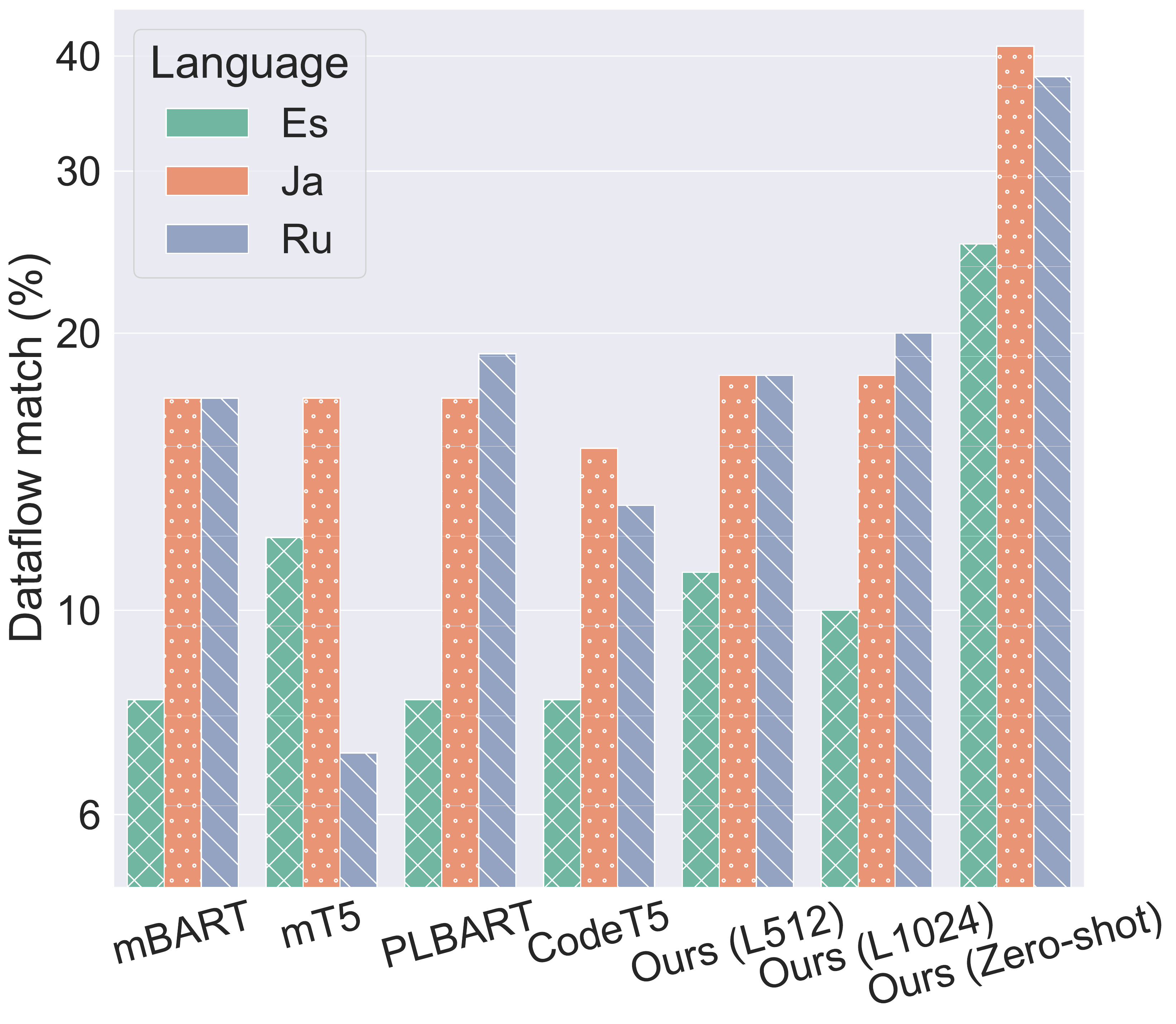}
         \vskip -1mm
         \caption{Semantic dataflow match (w/ log-scaled y-axis).} 
         \label{fig:dafaflow_bar}
     \end{subfigure}
     \hfill
     \begin{subfigure}[b]{0.46\textwidth}
         \centering
         \includegraphics[width=\textwidth]{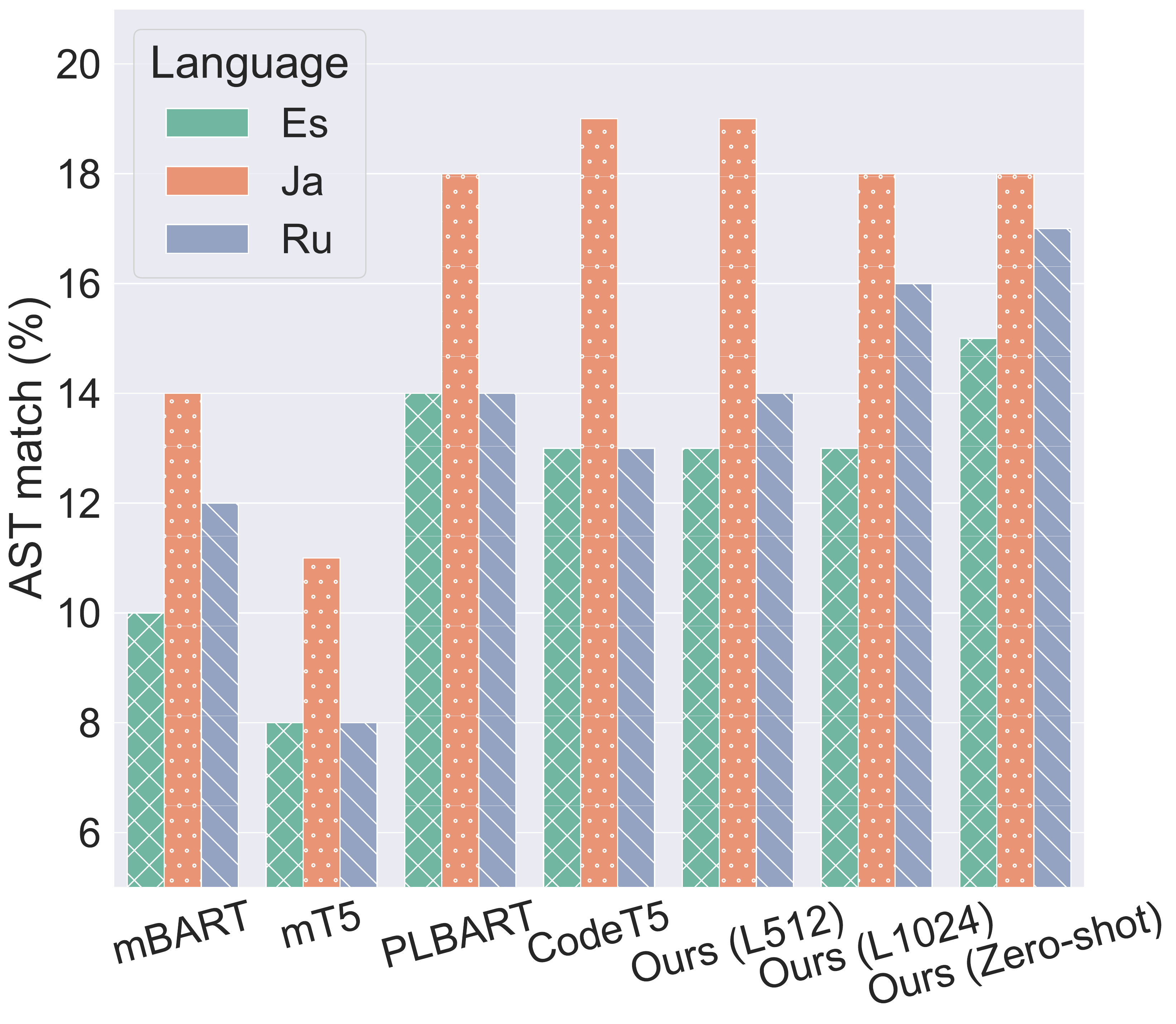} \vskip -2mm
         \caption{Syntactic AST match.} 
         \label{fig:AST_bar}
     \end{subfigure} \vskip -1mm
        \caption{Semantic and syntactic comparison on multilingual text-to-code generation. All comparison models are evaluated under ``translate-train'' settings by default, unless otherwise specified (\emph{i.e.}, ``zero-shot'').}
        \label{fig:codebleu_eval}
        \vskip -3mm
\end{figure*}

Figure~\ref{fig:codebleu_eval} illustrates the dataflow and AST match results of comparison models. PL baselines tend to generate code with better AST structures than NL models.
In particular, mT5 fails to produce code with proper AST syntax but can match or surpass others on dataflow evaluation except on Russian tasks. Our model (L512/1024) exceeds or matches baselines in terms of both the semantic dataflow and syntactic AST match.

\subsection{Ablation study}

\begin{figure}[h]
\vskip -1mm
\begin{center}
\includegraphics[width=\columnwidth]{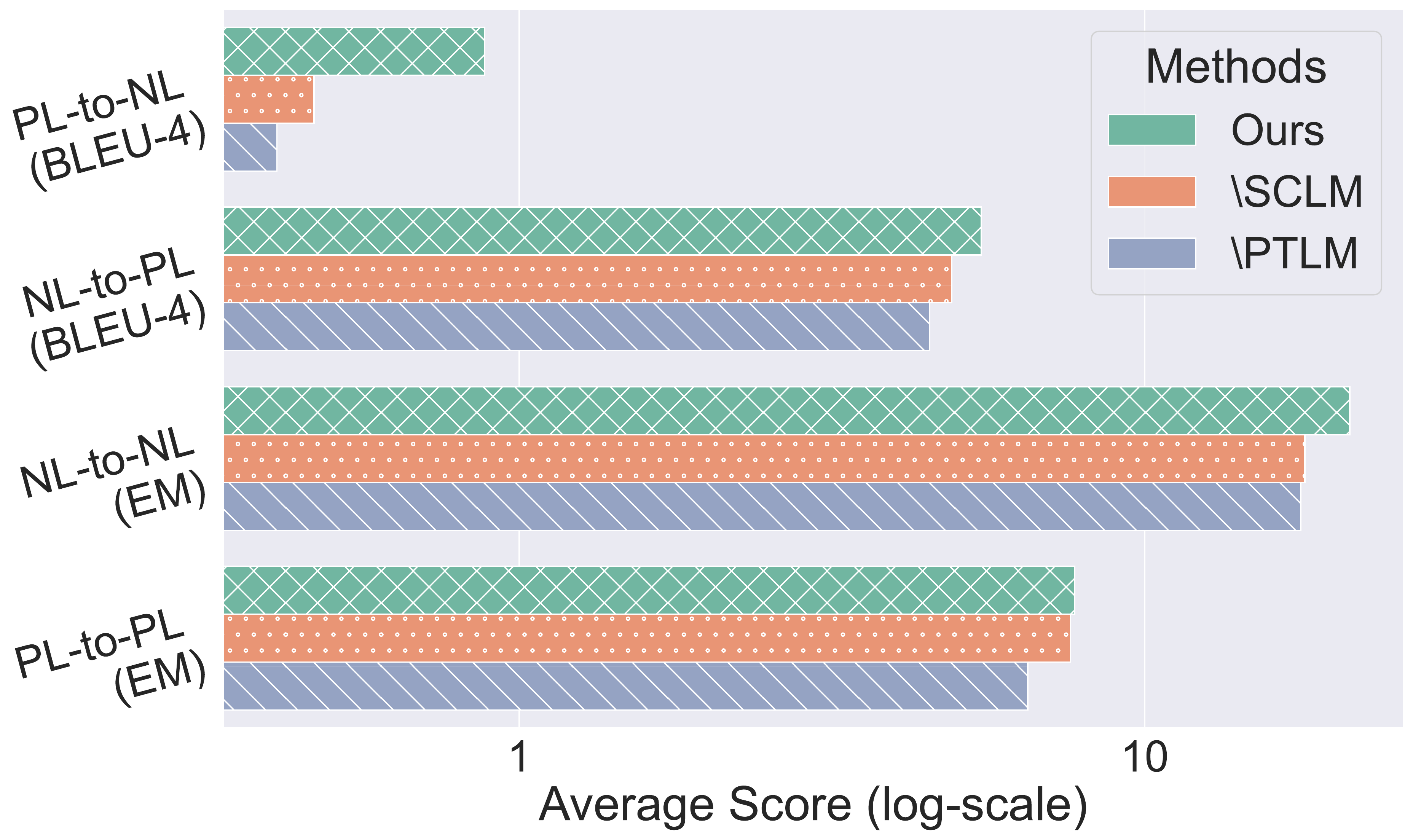}
\vskip -1mm
\caption{Ablation test performance (log-scale). The reported results are averaged among all subtasks.}
\vskip -2mm
\label{fig:ablation}
\end{center}
\end{figure}

\paragraph{Quantitative results} We carry out ablation experiments by ablating either SCLM or PTLM tasks and report the average results in Figure~\ref{fig:ablation}. It is shown that removing either monolingual ($\backslash$SCLM) or bilingual ($\backslash$PTLM) pre-training task could deteriorate overall performance of all tasks. Specifically, ablating PTLM would vastly reduce the performance of PL-to-NL and NL-to-PL tasks compared to removing SCLM, showing that pivot-based bi-text pre-training is crucial to implicit bridging between bilingual NL-to-PL or PL-to-NL pairs that never seen explicitly in training data. Meanwhile, PTLM contributes slightly more than SCLM in NL-to-NL translation. We suspect that although PLTM can provide explicit training on bilingual data, SCLM could implicitly learn NL patterns from amounts of monolingual training corpora.
In contrast, SCLM makes a trivial contribution to PL-to-PL generation, indicating that PTLM allows the model to focus on full-sequence generation instead of partial span reconstruction. Considering that the training data size of the PL corpus is quite limited, we suspect that pre-training on more open-source repositories from GitHub would bring more significant performance gain. We refer to \S\ref{ap:ablation} for detailed results on each subtask.

\begin{figure*}[h]
\begin{center}
\includegraphics[width=\textwidth]{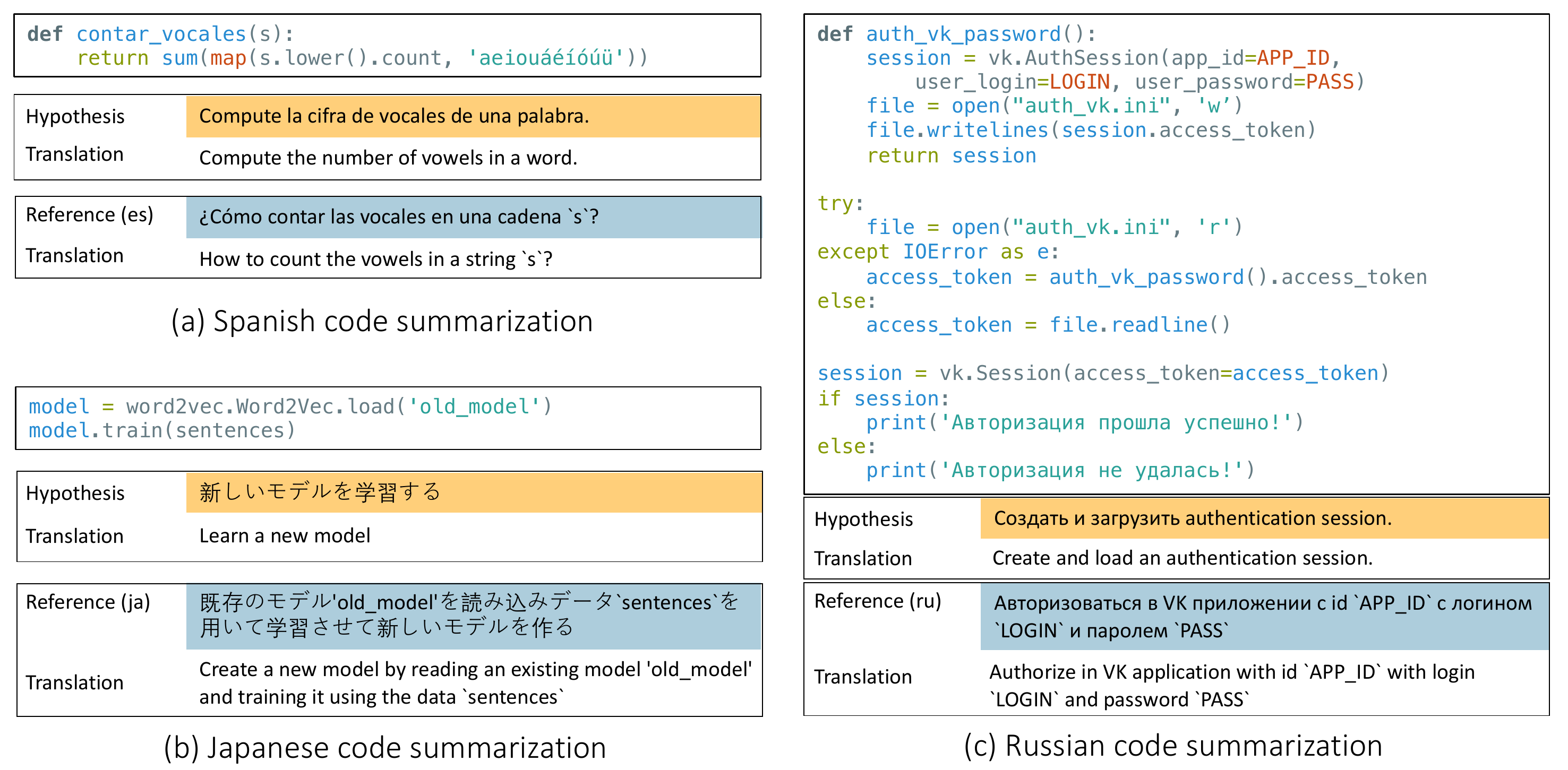}
\vskip -1mm
\caption{Examples of zero-shot multilingual code summarization (code-to-text). }
\label{fig:c2t-case}
\end{center}
\vskip -4mm
\end{figure*}

\begin{figure}[h]
\begin{center}
\includegraphics[width=\columnwidth]{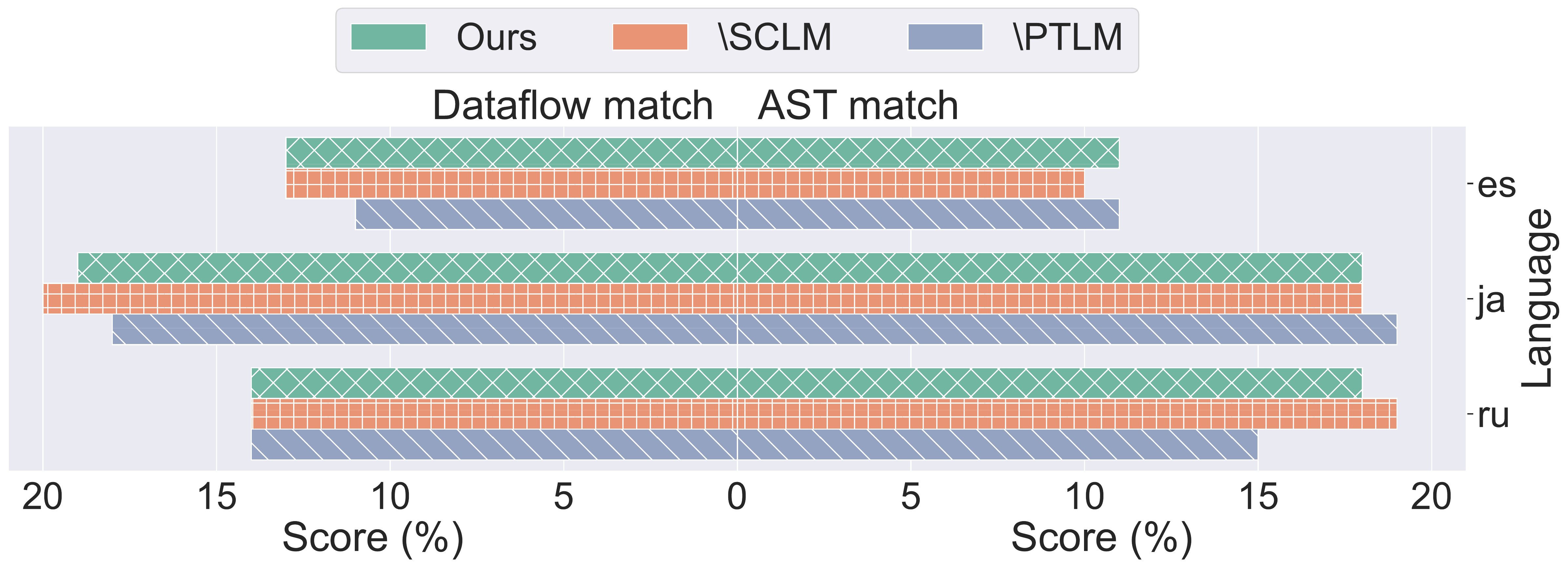}
\vskip -1mm
\caption{Ablation results on dataflow and AST match.}
\label{fig:ablation_codebleu}
\end{center}
\vskip -2mm
\end{figure}

\paragraph{Analyzing PL semantics \& syntax} We further analyze the semantic and syntactic structure of multilingual text-to-code generation for ablation comparison. Figure~\ref{fig:ablation_codebleu} shows dataflow and AST match performance on text-to-code generation given multilingual NL inputs. We find that removing SCLM does not overly impact the semantic dataflow and syntactic structures of generated PL. At the same time, ablating PTLM would generally cause more considerable fluctuation in the semantics and syntax of generated PL, suggesting that PTLM could allow the model to capture bilingual alignment and translation across multilingualism.
%


\subsection{Zero-shot prompting}
\label{sec:zero-shot}
To verify the zero-shot ability of ERNIE-Code, we carry out code-to-text, text-to-code, and text-to-text experiments with zero-shot prompting. Precisely, we prepend a prompt prefix ``translate \verb|S| to \verb|T|: \verb|\n|'' on the left of inputs, where \verb|S| and \verb|T| denote the source and target language respectively. Then we use beam search with five beams to obtain zero-shot predictions. 

\paragraph{Quantitative analysis} Table~\ref{tab:c2t} (last row) shows the performance of \textit{zero-shot} code-to-text generation. Our model demonstrates excellent zero-shot capability on Japanese and Russian summary generation, even outperforming ``translate-train'' settings by 0.43 / 9.05 on BLEU / ROUGE-L in general. This is because the training data is automatically translated rather than human annotated (\emph{i.e.}, ``translate-train'' settings), lowering the quality of training data. 
Table~\ref{tab:t2c} shows that our model can zero-shotly produce code fragments with higher CodeBLEU scores than ``translate-train'' settings. This indicates that our cross-lingual NL-PL pre-training renders excellent transfer learning capability in bridging multilingual NLs and PLs.


\paragraph{Zero-shot PL-to-NL generation} Figure~\ref{fig:c2t-case} exhibits \textit{zero-shot} multilingual code summarization examples in three target languages. Our model can attend to the whole picture of code semantics while ignoring blunt descriptions of detailed implementation, demonstrating the effectiveness of our approach on zero-shot prompting. To extend the evaluation to other NL, we further release a Python-Chinese test set by translating mCoNaLa into its Chinese variant via crowd-sourcing. Our model shows decent ability on zero-shot PL-to-Chinese generation.
We give zero-shot demonstrations and provide data curation details in \S\ref{ap:zh-test}. We argue that our model captures many NL genres via cross-lingual pre-training. We encourage the community to release more multilingual code-to-text benchmarks for further evaluation.

\paragraph{Qualitative examples (zero-shot)} We show a variety of qualitative examples with zero-shot prompting in \S\ref{ap:examples}: multilingual code summarization, NL-to-PL generation, zero-shot NL translation of technical jargon in eight randomly selected directions.



\section{Conclusion}
\label{sec:concl}
This work makes the first step towards explicitly connecting computer programs to human languages in a universal multilingual fashion. By virtue of cross-lingual pre-training on 116 NLs and 6 PLs, our model exhibits strong performance in various tasks across computer programs and natural languages, including PL-to-NL, NL-to-PL, NL-to-NL, and PL-to-PL. Our model shows descent zero-shot performance via prompting on PL summarization and NL translation. Finally, we provide discussions about limitations and future work for improvement.

\section*{Acknowledgements}
We would like to thank Xuhong Li and Qiwei Peng for their helpful feedback on the initial manuscript of this work.

\section*{Limitations}

\paragraph{Releasing multilingual NL-PL benchmark} While our model has been shown to capture multilingual languages between humans and computer programs, we could not systemically evaluate its performance on a wide range of multilingual NLs due to the lack of corresponding benchmarks. Instead, we undertake NL-to-PL and PL-to-NL experiments on mCoNaLa that involves only three NLs and present demonstration examples via zero-shot prompting to reveal its cross-lingual capacity. We encourage researchers in the community to release more multilingual NL-PL benchmarks to accelerate the development of this intersecting area. 

\paragraph{Scaling up the model size and data}
In this work, we only use the PL data from CodeSearchNet for a fair comparison to baselines, preventing the model from learning from more PL genres and billions of open-source repositories. Increasing the amount of data for bilingual NL-PL pairs is also a promising direction, such as using data augmentation. Moreover, the scaling law for large pre-training has been well studied and shown significant performance gains in the literature~\citep{Chen2021EvaluatingLL,Li2022CompetitionLevelCG}.
A targeted effort at expanding the pre-training data size and scaling up models could give rise to more considerable improvement toward universal multilingual NL-PL pre-training. 

\paragraph{Curse of multilinguality} We argue that the \emph{curse of multilinguality}~\citep{DBLP:conf/acl/ConneauKGCWGGOZ20} also exists in unified multilingual NL-PL pre-training, in which per-language capacity decreases as the number of languages increases given a fixed model size. It is an interesting direction to investigate the issue of \textit{curse of multilinguality} upon this work.



\bibliography{anthology,custom}
\bibliographystyle{acl_natbib}

\clearpage
\appendix
\section{Appendix}

\subsection{Input representation}
\label{ap:tokenization}
\begin{figure}[thb]
\begin{center}
\includegraphics[width=\columnwidth]{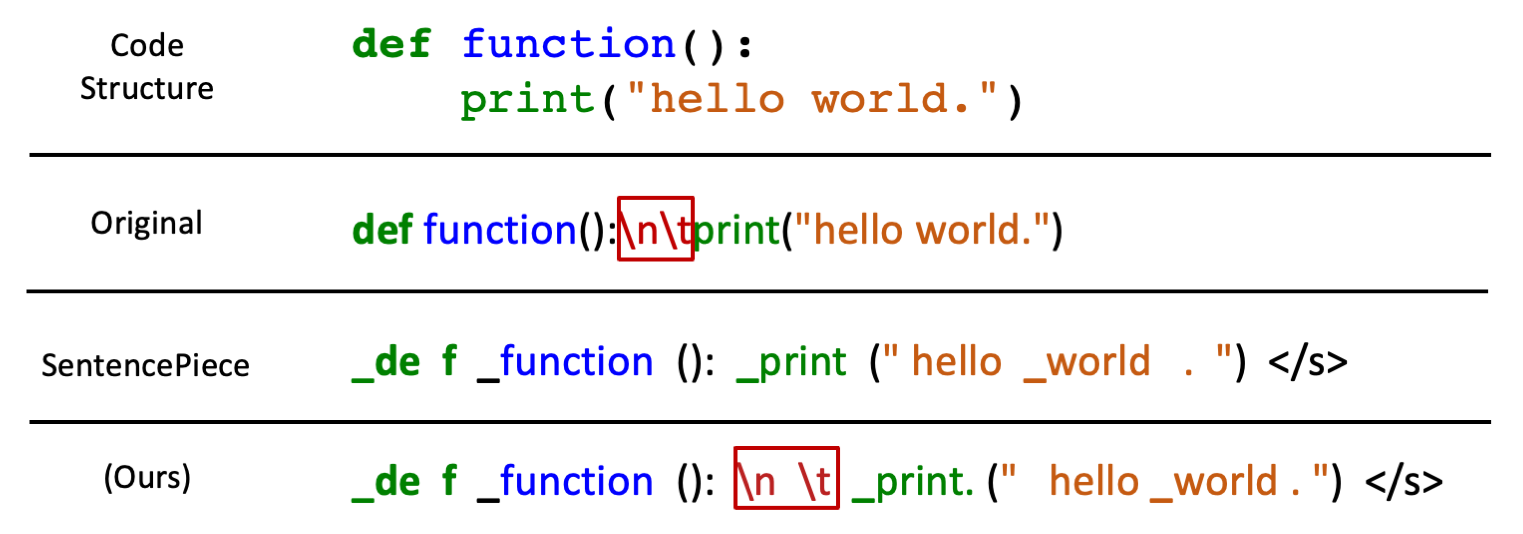}
\caption{NL/PL-shared tokenization example (Python). ``</s>'' represents the end-of-sentence token.}
\label{fig:tokenization}
\end{center}
\end{figure}

We base our shared text/code lexer on the mT5 tokenizer - SentencePiece~\citep{kudo-richardson-2018-sentencepiece}, specifically unigram language model~\citep{kudo-2018-subword}. Since the word distribution in PL essentially differs from that of NL, it is not feasible to directly apply the SentencePiece tokenization on PL. SentencePiece is ineffective in encoding whitespace characters - such as blank space, tab \verb|\t|, and newline character \verb|\n| - which are crucial in representing structures and indentations in source code. We thus add a set of additional tokens for encoding whitespace of different lengths in PL. Considering that developers with different programming habits may type indentations with various lengths and characters (tab or space), we add spaces of length-1/2/4 (denoted as  \verb|<space*1>|, \verb|<space*2>|, \verb|<space*4>|, respectively), and tab \verb|\t| to represent various indentations. Moreover, we use the newline symbol \verb|\n| to encode line breaks. Our tokenizer eventually consists of 250,105 SentencePiece vocabularies. Figure~\ref{fig:tokenization} exhibits a tokenization example of Python snippets. SentencePiece tends to normalize whitespaces and skip extra empty characters, while our modified tokenizer allows the model to cope with whitespace characters such as indentation in PL.

\subsection{Pre-training data}
\label{ap:data}
\subsubsection{PL data}
\label{ap:pl_data}
Table~\ref{tab:codesearchnet} shows the statistics of monolingual PL data and parallel NL-PL pairs, consisting of 6.5 million monolingual samples and 1.9 million NL-PL pairs in six different PLs. We do not use additional code repositories from GitHub for a fair comparison to baseline PL models. 

\begin{table}[thb]
\centering
\resizebox{.8\columnwidth}{!}{%
\begin{tabular}{@{}lrr@{}}
\toprule
\textbf{PL} & \textbf{\#Sample} & \multicolumn{1}{l}{\textbf{\#NL-PL pair}} \\ \midrule
Go          & 726,768                    & 317,832                                       \\
Java        & 1,569,889                  & 454,451                                       \\
JavaScript  & 1,857,835                  & 123,889                                       \\
PHP         & 977,821                    & 523,712                                       \\
Python      & 1,156,085                  & 412,178                                       \\
Ruby        & 164,048                    & 48,791                                        \\ \hline
\#Total     & 6,452,446                    & 1,880,853                                     \\ \bottomrule
\end{tabular}%
}
\caption{Statistics of CodeSearchNet in six PLs, totaling 6.5 million monolingual PL instances and 1.5 million parallel NL-PL samples.  }
\label{tab:codesearchnet}
\end{table}

The NL data may also exist in their paired PL data, serving as a comment or docstring. It could result in data leakage for PL-to-NL translation if NL has been given as a part of PL inputs, thereby hurting the code-to-text test performance. Accordingly, for all code-to-text generation and randomly 50\% of text-to-code generation in PTLM training, we replace all NL sentences as an NL placeholder ``<|removed|>'' if it exists in the corresponding PL fragments. 

We additionally observe that parallel data in CodeSearchNet only contain few non-English NLs. Directly regarding all NLs in CodeSearchNet as English would confuse the model to distinguish various NLs. To better leverage this parallel supervision signal, we utilize FastText~\citep{joulin2016bag} tools\footnote{\url{https://fasttext.cc/docs/en/language-identification.html}} to identify different NLs. Specifically, we only consider NL sentences with confidence higher than 80\% predicted by FastText. In PTLM training, we use the predicted language genre with 50\% probability at random; otherwise, we treat the sample as ``text'' other than ``English''. Therefore, the model could implicitly tell different language genres without being exposed to erroneous supervision.

\subsubsection{NL data}
\label{ap:nl_data}
\paragraph{Monolingual NL corpus} CC-100\footnote{\url{https://data.statmt.org/cc-100}} was constructed by processing CommonCrawl snapshots~\citep{Wenzek2019CCNetEH}. The original CC-100 dataset comprises documents separated by double newlines. We maintain the document-level corpus by concatenating paragraphs within the same document page. Table~\ref{tab:CC100} summarizes the statistics of our processed data, totaling 1.5 billion training document pages in 116 monolingual NLs. We rescale the data distribution according to page counts as aforementioned in Eq.~\eqref{eq:rescale} with $\alpha=0.3$.

\begin{table*}[]
\centering
\resizebox{.85\textwidth}{!}{%
\begin{tabular}{@{}cccr|cccr@{}}
\toprule
\textbf{\begin{tabular}[c]{@{}c@{}}ISO \\ code\end{tabular}} & \textbf{Language} & \textbf{\begin{tabular}[c]{@{}c@{}}\#Pages\\ (M)\end{tabular}} & \multicolumn{1}{c|}{\textbf{\begin{tabular}[c]{@{}c@{}}Percent.\\ (\%)\end{tabular}}} & \textbf{\begin{tabular}[c]{@{}c@{}}ISO \\ code\end{tabular}} & \textbf{Language} & \textbf{\begin{tabular}[c]{@{}c@{}}\#Pages\\ (M)\end{tabular}} & \multicolumn{1}{c}{\textbf{\begin{tabular}[c]{@{}c@{}}Percent.\\ (\%)\end{tabular}}} \\ \midrule
af & Afrikaans & 1.3 & 0.09 & lt & Lithuanian & 9.19 & 0.61 \\
am & Amharic & 0.24 & 0.02 & lv & Latvian & 5.83 & 0.39 \\
ar & Arabic & 15.04 & 1.0 & mg & Malagasy & 0.15 & 0.01 \\
as & Assamese & 0.05 & 0.0 & mk & Macedonian & 1.78 & 0.12 \\
az & Azerbaijani & 4.1 & 0.27 & ml & Malayalam & 1.9 & 0.13 \\
be & Belarusian & 1.45 & 0.1 & mn & Mongolian & 0.96 & 0.06 \\
bg & Bulgarian & 18.16 & 1.21 & mr & Marathi & 1.01 & 0.07 \\
bn & Bengali & 4.11 & 0.27 & ms & Malay & 11.92 & 0.79 \\
bn\_rom & Bengali Romanized & 6.5 & 0.43 & my & Burmese & 0.22 & 0.01 \\
br & Breton & 0.14 & 0.01 & my\_zaw & Burmese (Zawgyi) & 0.88 & 0.06 \\
bs & Bosnian & 0.4 & 0.03 & ne & Nepali & 1.13 & 0.08 \\
ca & Catalan & 7.01 & 0.47 & nl & Dutch & 31.16 & 2.08 \\
cs & Czech & 10.15 & 0.68 & no & Norwegian & 28.8 & 1.92 \\
cy & Welsh & 0.71 & 0.05 & ns & Northern Sotho & 0.03 & 0.0 \\
da & Danish & 30.19 & 2.01 & om & Oromo & 0.08 & 0.01 \\
de & German & 69.02 & 4.6 & or & Oriya & 0.19 & 0.01 \\
el & \begin{tabular}[c]{@{}c@{}}Modern Greek \end{tabular} & 12.33 & 0.82 & pa & Panjabi & 0.33 & 0.02 \\
en & English & 247.59 & 16.49 & pl & Polish & 31.2 & 2.08 \\
eo & Esperanto & 0.58 & 0.04 & ps & Pushto & 0.26 & 0.02 \\
es & Spanish & 60.54 & 4.03 & pt & Portuguese & 39.0 & 2.6 \\
et & Estonian & 3.94 & 0.26 & qu & Quechua & 0.03 & 0.0 \\
eu & Basque & 1.86 & 0.12 & rm & Romansh & 0.03 & 0.0 \\
fa & Persian & 36.96 & 2.46 & ro & Romanian & 30.21 & 2.01 \\
ff & Fulah & 0.02 & 0.0 & ru & Russian & 123.18 & 8.2 \\
fi & Finnish & 28.12 & 1.87 & sa & Sanskrit & 0.12 & 0.01 \\
fr & French & 62.11 & 4.14 & sc & Sardinian & 0.0 & 0.0 \\
fy & Western Frisian & 0.2 & 0.01 & sd & Sindhi & 0.08 & 0.01 \\
ga & Irish & 0.52 & 0.03 & si & Sinhala & 0.67 & 0.04 \\
gd & Scottish Gaelic & 0.11 & 0.01 & sk & Slovak & 17.0 & 1.13 \\
gl & Galician & 1.85 & 0.12 & sl & Slovenian & 6.24 & 0.42 \\
gn & Guarani & 0.02 & 0.0 & so & Somali & 0.4 & 0.03 \\
gu & Gujarati & 0.75 & 0.05 & sq & Albanian & 2.72 & 0.18 \\
ha & Hausa & 0.46 & 0.03 & sr & Serbian & 2.7 & 0.18 \\
he & Hebrew & 12.77 & 0.85 & ss & Swati & 0.0 & 0.0 \\
hi & Hindi & 8.11 & 0.54 & su & Sundanese & 0.06 & 0.0 \\
hi\_rom & Hindi Romanized & 1.97 & 0.13 & sv & Swedish & 46.77 & 3.12 \\
hr & Croatian & 16.54 & 1.1 & sw & Swahili & 1.13 & 0.08 \\
ht & Haitian & 0.09 & 0.01 & ta & Tamil & 4.12 & 0.27 \\
hu & Hungarian & 26.14 & 1.74 & ta\_rom & Tamil Romanized & 1.6 & 0.11 \\
hy & Armenian & 2.14 & 0.14 & te & Telugu & 1.21 & 0.08 \\
id & Indonesian & 79.68 & 5.31 & te\_rom & Telugu Romanized & 1.9 & 0.13 \\
ig & Igbo & 0.04 & 0.0 & th & Thai & 23.92 & 1.59 \\
is & Icelandic & 2.06 & 0.14 & tl & Tagalog & 2.64 & 0.18 \\
it & Italian & 24.67 & 1.64 & tn & Tswana & 0.24 & 0.02 \\
ja & Japanese & 65.61 & 4.37 & tr & Turkish & 18.42 & 1.23 \\
jv & Javanese & 0.31 & 0.02 & ug & Uighur & 0.11 & 0.01 \\
ka & Georgian & 2.68 & 0.18 & uk & Ukrainian & 24.98 & 1.66 \\
kk & Kazakh & 1.77 & 0.12 & ur & Urdu & 2.26 & 0.15 \\
km & Central Khmer & 0.61 & 0.04 & ur\_rom & Urdu Romanized & 4.58 & 0.3 \\
kn & Kannada & 0.91 & 0.06 & uz & Uzbek & 0.46 & 0.03 \\
ko & Korean & 35.68 & 2.38 & vi & Vietnamese & 52.48 & 3.5 \\
ku & Kurdish & 0.24 & 0.02 & wo & Wolof & 0.13 & 0.01 \\
ky & Kirghiz & 0.41 & 0.03 & xh & Xhosa & 0.15 & 0.01 \\
la & Latin & 3.1 & 0.21 & yi & Yiddish & 0.15 & 0.01 \\
lg & Ganda & 0.09 & 0.01 & yo & Yoruba & 0.02 & 0.0 \\
li & Limburgan & 0.02 & 0.0 & zh & Chinese (Simplified) & 40.0 & 2.66 \\
ln & Lingala & 0.02 & 0.0 & zh-Hant & Chinese (Traditional) & 12.33 & 0.82 \\
lo & Lao & 0.2 & 0.01 & zu & Zulu & 0.07 & 0.0 \\ \bottomrule
\end{tabular}%
}
\caption{Statistics of CC-100 corpus, totaling 1.5 billion training document pages from 116 different NLs. Reported training pages and percentages are calculated according to the document distribution of original data. Note that our 116 NLs include 5 Romanized variants of existing languages denoted by ``Romanized''.}
\label{tab:CC100}
\end{table*}

\begin{table*}[]
\centering
\resizebox{\textwidth}{!}{%
\begin{tabular}{@{}cccccccccc@{}}
\toprule
\textbf{\begin{tabular}[c]{@{}c@{}}ISO \\ code\end{tabular}} & \textbf{Lang 1} & \textbf{Lang 2} & \textbf{\begin{tabular}[c]{@{}c@{}}\#Pairs\\ (M)\end{tabular}} & \multicolumn{1}{c|}{\textbf{\begin{tabular}[c]{@{}c@{}}Percent.\\ (\%)\end{tabular}}} & \textbf{\begin{tabular}[c]{@{}c@{}}ISO \\ code\end{tabular}} & \textbf{Language 1} & \textbf{Language 2} & \textbf{\begin{tabular}[c]{@{}c@{}}\#Pairs\\ (M)\end{tabular}} & \textbf{\begin{tabular}[c]{@{}c@{}}Percent.\\ (\%)\end{tabular}} \\ \midrule
ar-bg & Arabic & Bulgarian & 46.57 & \multicolumn{1}{c|}{0.59} & en-ru & English & Russian & 312.91 & 3.99 \\
ar-de & Arabic & German & 44.58 & \multicolumn{1}{c|}{0.57} & en-sw & English & Swahili & 9.41 & 0.12 \\
ar-el & Arabic & Greek & 45.66 & \multicolumn{1}{c|}{0.58} & en-th & English & Thai & 26.11 & 0.33 \\
ar-en & Arabic & English & 199.26 & \multicolumn{1}{c|}{2.54} & en-tr & English & Turkish & 196.96 & 2.51 \\
ar-es & Arabic & Spanish & 141.9 & \multicolumn{1}{c|}{1.81} & en-ur & English & Urdu & 11.04 & 0.14 \\
ar-fr & Arabic & French & 118.52 & \multicolumn{1}{c|}{1.51} & en-vi & English & Vietnamese & 79.56 & 1.02 \\
ar-hi & Arabic & Hindi & 7.24 & \multicolumn{1}{c|}{0.09} & en-zh & English & Chinese & 156.31 & 1.99 \\
ar-ru & Arabic & Russian & 96.15 & \multicolumn{1}{c|}{1.23} & es-fr & Spanish & French & 522.47 & 6.67 \\
ar-sw & Arabic & Swahili & 2.38 & \multicolumn{1}{c|}{0.03} & es-hi & Spanish & Hindi & 15.93 & 0.2 \\
ar-th & Arabic & Thai & 9.42 & \multicolumn{1}{c|}{0.12} & es-ru & Spanish & Russian & 166.12 & 2.12 \\
ar-tr & Arabic & Turkish & 58.32 & \multicolumn{1}{c|}{0.74} & es-sw & Spanish & Swahili & 7.88 & 0.1 \\
ar-ur & Arabic & Urdu & 2.43 & \multicolumn{1}{c|}{0.03} & es-th & Spanish & Thai & 10.15 & 0.13 \\
ar-vi & Arabic & Vietnamese & 17.36 & \multicolumn{1}{c|}{0.22} & es-tr & Spanish & Turkish & 105.87 & 1.35 \\
ar-zh & Arabic & Chinese & 55.68 & \multicolumn{1}{c|}{0.71} & es-ur & Spanish & Urdu & 0.8 & 0.01 \\
bg-de & Bulgarian & German & 57.71 & \multicolumn{1}{c|}{0.74} & es-vi & Spanish & Vietnamese & 44.33 & 0.57 \\
bg-el & Bulgarian & Greek & 68.07 & \multicolumn{1}{c|}{0.87} & es-zh & Spanish & Chinese & 74.93 & 0.96 \\
bg-en & Bulgarian & English & 151.04 & \multicolumn{1}{c|}{1.93} & fr-hi & French & Hindi & 15.38 & 0.2 \\
bg-es & Bulgarian & Spanish & 86.31 & \multicolumn{1}{c|}{1.1} & fr-ru & French & Russian & 154.58 & 1.97 \\
bg-fr & Bulgarian & French & 69.09 & \multicolumn{1}{c|}{0.88} & fr-sw & French & Swahili & 8.91 & 0.11 \\
bg-hi & Bulgarian & Hindi & 3.35 & \multicolumn{1}{c|}{0.04} & fr-th & French & Thai & 8.7 & 0.11 \\
bg-ru & Bulgarian & Russian & 66.25 & \multicolumn{1}{c|}{0.85} & fr-tr & French & Turkish & 85.83 & 1.1 \\
bg-sw & Bulgarian & Swahili & 1.12 & \multicolumn{1}{c|}{0.01} & fr-ur & French & Urdu & 0.74 & 0.01 \\
bg-th & Bulgarian & Thai & 6.98 & \multicolumn{1}{c|}{0.09} & fr-vi & French & Vietnamese & 25.37 & 0.32 \\
bg-tr & Bulgarian & Turkish & 66.06 & \multicolumn{1}{c|}{0.84} & fr-zh & French & Chinese & 70.14 & 0.9 \\
bg-ur & Bulgarian & Urdu & 0.59 & \multicolumn{1}{c|}{0.01} & hi-ru & Hindi & Russian & 7.32 & 0.09 \\
bg-vi & Bulgarian & Vietnamese & 11.23 & \multicolumn{1}{c|}{0.14} & hi-sw & Hindi & Swahili & 1.46 & 0.02 \\
bg-zh & Bulgarian & Chinese & 11.56 & \multicolumn{1}{c|}{0.15} & hi-th & Hindi & Thai & 2.69 & 0.03 \\
de-el & German & Greek & 72.85 & \multicolumn{1}{c|}{0.93} & hi-tr & Hindi & Turkish & 8.75 & 0.11 \\
de-en & German & English & 655.83 & \multicolumn{1}{c|}{8.37} & hi-ur & Hindi & Urdu & 1.49 & 0.02 \\
de-es & German & Spanish & 242.73 & \multicolumn{1}{c|}{3.1} & hi-vi & Hindi & Vietnamese & 6.11 & 0.08 \\
de-fr & German & French & 269.02 & \multicolumn{1}{c|}{3.43} & hi-zh & Hindi & Chinese & 2.39 & 0.03 \\
de-hi & German & Hindi & 9.36 & \multicolumn{1}{c|}{0.12} & ru-sw & Russian & Swahili & 2.17 & 0.03 \\
de-ru & German & Russian & 80.08 & \multicolumn{1}{c|}{1.02} & ru-th & Russian & Thai & 8.12 & 0.1 \\
de-sw & German & Swahili & 3.22 & \multicolumn{1}{c|}{0.04} & ru-tr & Russian & Turkish & 51.77 & 0.66 \\
de-th & German & Thai & 7.07 & \multicolumn{1}{c|}{0.09} & ru-ur & Russian & Urdu & 2.56 & 0.03 \\
de-tr & German & Turkish & 57.14 & \multicolumn{1}{c|}{0.73} & ru-vi & Russian & Vietnamese & 16.47 & 0.21 \\
de-ur & German & Urdu & 0.86 & \multicolumn{1}{c|}{0.01} & ru-zh & Russian & Chinese & 61.53 & 0.79 \\
de-vi & German & Vietnamese & 20.77 & \multicolumn{1}{c|}{0.27} & sw-th & Swahili & Thai & 0.49 & 0.01 \\
de-zh & German & Chinese & 22.8 & \multicolumn{1}{c|}{0.29} & sw-tr & Swahili & Turkish & 4.16 & 0.05 \\
el-en & Greek & English & 190.87 & \multicolumn{1}{c|}{2.44} & sw-ur & Swahili & Urdu & 0.39 & 0.0 \\
el-es & Greek & Spanish & 133.05 & \multicolumn{1}{c|}{1.7} & sw-vi & Swahili & Vietnamese & 3.02 & 0.04 \\
el-fr & Greek & French & 117.73 & \multicolumn{1}{c|}{1.5} & sw-zh & Swahili & Chinese & 1.08 & 0.01 \\
el-hi & Greek & Hindi & 4.55 & \multicolumn{1}{c|}{0.06} & th-tr & Thai & Turkish & 9.26 & 0.12 \\
el-ru & Greek & Russian & 45.1 & \multicolumn{1}{c|}{0.58} & th-ur & Thai & Urdu & 0.64 & 0.01 \\
el-sw & Greek & Swahili & 1.84 & \multicolumn{1}{c|}{0.02} & th-vi & Thai & Vietnamese & 4.62 & 0.06 \\
el-th & Greek & Thai & 5.83 & \multicolumn{1}{c|}{0.07} & th-zh & Thai & Chinese & 0.97 & 0.01 \\
el-tr & Greek & Turkish & 69.81 & \multicolumn{1}{c|}{0.89} & tr-ur & Turkish & Urdu & 4.34 & 0.06 \\
el-ur & Greek & Urdu & 0.31 & \multicolumn{1}{c|}{0.0} & tr-vi & Turkish & Vietnamese & 16.29 & 0.21 \\
el-vi & Greek & Vietnamese & 14.84 & \multicolumn{1}{c|}{0.19} & tr-zh & Turkish & Chinese & 14.62 & 0.19 \\
el-zh & Greek & Chinese & 11.44 & \multicolumn{1}{c|}{0.15} & ur-vi & Urdu & Vietnamese & 0.58 & 0.01 \\
en-es & English & Spanish & 1088.62 & \multicolumn{1}{c|}{13.89} & ur-zh & Urdu & Chinese & 0.11 & 0.0 \\
en-fr & English & French & 884.16 & \multicolumn{1}{c|}{11.28} & vi-zh & Vietnamese & Chinese & 9.31 & 0.12 \\
en-hi & English & Hindi & 27.42 & 0.35 &  &  &  &  &  \\ \bottomrule
\end{tabular}%
}
\caption{Statistics of OPUS corpus, totaling 7.8 billion bilingual NL pairs from 105 different NL pairs. The reported count of bilingual pairs (``\#Sent.'') and percentage (``\#Percent.'') are calculated according to the original data. }
\label{tab:OPUS}
\end{table*}

\paragraph{Parallel NL corpus} We use parallel NL data collected from OPUS website\footnote{\url{https://opus.nlpl.eu}}. We summarize the statistics of collected OPUS data in Table~\ref{tab:OPUS}. The data we use are in 15 different NLs, comprising of 105 various bilingual language pairs (ignoring the dual direction between two languages) and 7.8 billion sentence pairs in total. Similar to CC-100 preprocessing, we apply the same data resampling process by following Eq.~\eqref{eq:rescale}, with $\alpha=0.3$.

 \subsubsection{Data rebalance between NL and PL} Considering that the data amount of PL and NL data vastly differs, the data distribution across NL and PL will still be unbalanced even after rescaling as per Eq.~\eqref{eq:rescale}, which could give rise in biases towards high-resource modality (\emph{i.e.}, NL). To mitigate this issue, we set the data distribution of PL and NL as 1:1 by equating the training sample ratio of PL with that of NL during pre-training. In other words, we train the same sample counts for NL and PL corpora.

\subsection{Experimental settings}
\label{ap:settings}

\subsubsection{Pre-training settings}
\label{ap:pretrain}

We use the same T5 architecture with a 12-layer encoder, a 12-layer decoder, 768 hidden units ($d_\text{model}$), 12 heads, 2048 feedforward linear units ($d_\text{ff}$), GELU activations, a dropout~\citep{srivastava2014dropout} rate as 0.1, and no embedding tying. \citet{Chen2021EvaluatingLL} find no difference between training from pre-trained model weights and that from scratch, except that the former converges more quickly. To this end, we use mT5 checkpoint\footnote{\url{https://github.com/google-research/multilingual-t5\#released-model-checkpoints}} for initialization, which already contains strong multilingual NL representations.

For pre-training, we set the maximum length (L) of 512/1024, a micro-batch size of 8/4 with a gradient accumulation step of 15. We utilize the Adafactor~\citep{shazeer2018adafactor} optimizer and a linear warmup of 1000 steps with a peak learning rate of 1e-4. All pre-training tasks are run on a cluster of 32 NVIDIA A100 GPUs with 40G memory for 100,000 training steps. To accelerate the pre-training, we utilize the ZeRO stage1 approach~\citep{rajbhandari2020zero} for partitioning optimizer states and enable BFloat16 half-precision format for mixed-precision training. The total pre-training time lasts around four weeks.

\subsubsection{Evaluation datasets}
Table~\ref{tab:datasets} reports the detailed statistics of evaluation dataset across a suit of code benchmarks, including NL-to-PL, PL-to-NL, PL-to-PL, and NL-to-NL.

\begin{table}[thb]
\centering
\resizebox{\columnwidth}{!}{%
\begin{tabular}{@{}lllrrr@{}}
\toprule
Task & Dataset & Language & \multicolumn{1}{c}{Train} & \multicolumn{1}{c}{Valid} & \multicolumn{1}{c}{Test} \\ \midrule
\multirow{3}{*}{NL-PL} & \multirow{3}{*}{\begin{tabular}[c]{@{}l@{}}mCoNaLa\\ \citep{Wang2022MCoNaLaAB}\end{tabular}} & Spanish $\leftrightarrow$ Python & - & - & 341 \\ \cmidrule(l){3-6} 
 &  & Japanese $\leftrightarrow$ Python & - & - & 210 \\ \cmidrule(l){3-6} 
 &  & Russian $\leftrightarrow$ Python & - & - & 345 \\ \midrule
\multirow{2}{*}{PL-PL} & \multirow{2}{*}{\begin{tabular}[c]{@{}l@{}}Bugs2Fix\\ \citep{Tufano2019AnES}\end{tabular}} & Java-small & 46,680 & 5,835 & 5,835 \\ \cmidrule(l){3-6} 
 &  & Java-medium & 52,364 & 6,545 & 6,545 \\ \midrule
\multirow{4}{*}{\begin{tabular}[c]{@{}l@{}}\\NL-NL\end{tabular} } & \multirow{4}{*}{\begin{tabular}[c]{@{}l@{}}\\Microsoft Docs\\ \citep{Lu2021CodeXGLUEAM}\end{tabular}} & Danish$\leftrightarrow$English & 42,701 & 1,000 & 1,000 \\ \cmidrule(l){3-6} 
 &  & Latvian$\leftrightarrow$English & 18,749 & 1,000 & 1,000 \\ \cmidrule(l){3-6} 
 &  & Norwegian$\leftrightarrow$English & 44,322 & 1,000 & 1,000 \\ \cmidrule(l){3-6} 
 &  & Chinese$\leftrightarrow$English & 50,154 & 1,000 & 1,000 \\ \bottomrule
\end{tabular}%
}
\caption{Statistics of downstream benchmark datasets.}
\label{tab:datasets}
\end{table}

\subsubsection{Finetuning settings}
\label{ap:finetune} 
When finetuning on end tasks, we use mini-batches of 8/4, and a maximum input length of 512. We set the maximum target length as 128, 64, 256, and 256 for code summarization, text-to-code, documentation translation, and code repair tasks, respectively. We use prompt-based finetuning by prepending a task prompt (as shown in Table~\ref{tab:prompt}) before each sample for training and evaluation. We finetune code-to-text, text-to-code, and documentation translation tasks for 100 epochs and train 10 epochs on the code repair dataset. 
For all finetuning experiments, we use the AdamW~\citep{Loshchilov2019DecoupledWD} optimizer with a learning rate of 5e-5. As to model inference, we apply beam search decoding with five beams. We conducted all finetuning experiments on 8 NVIDIA V100 GPUs with 32G memory.

\begin{table}[]
\centering
\resizebox{\columnwidth}{!}{%
\begin{tabular}{@{}cl@{}}
\toprule
\multicolumn{1}{c}{\textbf{Finetuning task}} & \multicolumn{1}{c}{\textbf{Prompt format}} \\ \midrule
\multirow{3}{*}{\textbf{Code-to-text}} & ``translate Spanish to Python: $\backslash$n'' \\
 & ``translate Japanese to Python: $\backslash$n'' \\
 & ``translate Russian to Python: $\backslash$n'' \\ \midrule
\multirow{3}{*}{\textbf{Text-to-code}} & ``translate Python to Spanish: $\backslash$n'' \\
 & ``translate Python to Japanese: $\backslash$n'' \\
 & ``translate Python to Russian: $\backslash$n'' \\ \midrule
\begin{tabular}[c]{@{}c@{}}\textbf{Documentation}\\ \textbf{translation}\end{tabular} & ``translate `src\_lang' to `tgt\_lang':$\backslash$n'' \\ \midrule
\textbf{Code repair} & ``fix bugs: $\backslash$n'' \\ \bottomrule
\end{tabular}%
}
\caption{Task prompt we use for finetuning. For documentation translation, the ``src\_lang'' and ``tgt\_lang'' represent the source and target language (\emph{e.g.}, English, Danish, Latvian, Norwegian, and Chinese), respectively.}
\label{tab:prompt}
\end{table}

\subsection{Ablation results}
\label{ap:ablation}
Table~\ref{tab:ablation} reports ablation results of  code summarization (mCoNaLa), text-to-code generation (mCoNaLa), documentation translation (Microsoft Docs), and code repair (Bugs2Fix), showing that combining SCLM and PTLM can confer benefit for all of the end tasks.

\begin{table*}[h]
    \begin{subtable}[h]{\textwidth}
    \centering
    \resizebox{\textwidth}{!}{%
    \begin{tabular}{c|rrr|rrr|rrr|rrr}
    \hline
    \multirow{2}{*}{\textbf{Model}} & \multicolumn{3}{c|}{\textbf{es}} & \multicolumn{3}{c|}{\textbf{ja}} & \multicolumn{3}{c|}{\textbf{ru}} & \multicolumn{3}{c}{\textbf{Avg.}} \\ \cline{2-13} 
     & \multicolumn{1}{l}{\textbf{B-4}} & \multicolumn{1}{l}{\textbf{R-L}} & \multicolumn{1}{l|}{\textbf{chrF}} & \multicolumn{1}{l}{\textbf{B-4}} & \multicolumn{1}{l}{\textbf{R-L}} & \multicolumn{1}{l|}{\textbf{chrF}} & \multicolumn{1}{l}{\textbf{B-4}} & \multicolumn{1}{l}{\textbf{R-L}} & \multicolumn{1}{l|}{\textbf{chrF}} & \multicolumn{1}{l}{\textbf{B-4}} & \multicolumn{1}{l}{\textbf{R-L}} & \multicolumn{1}{l}{\textbf{chrF}} \\ \hline
    \multicolumn{1}{l|}{\textbf{Ours}} & 1.90 & 32.51 & 23.22 & 0.30 & 10.62 & 9.16 & 0.43 & 5.01 & 16.60 & 0.88 & 16.05 & 16.33 \\ \hdashline
    $\backslash$\textbf{SCLM} & 1.04 & 23.96 & 19.56 & 0.17 & 7.62 & 8.88 & 0.21 & 2.69 & 15.53 & 0.47 & 11.42 & 15.10 \\
    $\backslash$\textbf{PTLM} & 0.96 & 22.47 & 24.00 & 0.06 & 5.71 & 8.22 & 0.20 & 4.92 & 14.66 & 0.41 & 11.03 & 14.15 \\ \hline
    \end{tabular}%
    }
\caption{Ablation results on multilingual code summarization.}
\label{tab:c2t-ablation}
    \end{subtable}
    \hfill
    \begin{subtable}[h]{\textwidth}
    \centering
      \resizebox{\textwidth}{!}{%
    \begin{tabular}{c|lll|lll|lll|lll}
\hline
\multirow{2}{*}{\textbf{Model}} & \multicolumn{3}{c|}{\textbf{es}} & \multicolumn{3}{c|}{\textbf{ja}} & \multicolumn{3}{c|}{\textbf{ru}} & \multicolumn{3}{c}{\textbf{Avg.}} \\ \cline{2-13} 
 & \textbf{B-4} & \textbf{R-L} & \textbf{C-B} & \textbf{B-4} & \textbf{R-L} & \textbf{C-B} & \textbf{B-4} & \textbf{R-L} & \textbf{C-B} & \textbf{B-4} & \textbf{R-L} & \textbf{C-B} \\ \hline
\multicolumn{1}{l|}{\textbf{Ours}} & \multicolumn{1}{r}{2.25} & \multicolumn{1}{r}{14.92} & \multicolumn{1}{r|}{0.06} & 8.06 & 22.65 & 0.10 & 6.12 & 25.27 & 0.08 & 5.48 & 20.95 & 0.08 \\ \hdashline
$\backslash$\textbf{SCLM} & 2.42 & 14.27 & 0.06 & 6.89 & 21.31 & 0.10 & 5.41 & 23.09 & 0.08 & 4.91 & 19.56 & 0.08 \\
$\backslash$\textbf{PTLM} & 2.08 & 13.94 & 0.06 & 6.40 & 17.77 & 0.10 & 5.11 & 23.17 & 0.08 & 4.53 & 18.29 & 0.08 \\ \hline
\end{tabular}%
    }
\caption{Ablation results on multilingual text-to-code generation.}
\label{tab:t2c-ablation}
     \end{subtable}
     \hfill
    \begin{subtable}[h]{\textwidth}
        \centering
       \resizebox{\textwidth}{!}{%
\begin{tabular}{l|crrr|crrr|crrr|crrr|ll}
\hline
\multirow{3}{*}{\textbf{Model}} & \multicolumn{4}{c|}{\textbf{En-Da}} & \multicolumn{4}{c|}{\textbf{En-Lv}} & \multicolumn{4}{c|}{\textbf{En-No}} & \multicolumn{4}{c|}{\textbf{En-Zh}} & \multirow{3}{*}{\textbf{\begin{tabular}[c]{@{}l@{}}Avg.\\ B-4\end{tabular}}} & \multirow{3}{*}{\textbf{\begin{tabular}[c]{@{}l@{}}Avg.\\ EM\end{tabular}}} \\ \cline{2-17}
 & \multicolumn{2}{c|}{$\rightarrow$} & \multicolumn{2}{c|}{$\leftarrow$} & \multicolumn{2}{c|}{$\rightarrow$} & \multicolumn{2}{c|}{$\leftarrow$} & \multicolumn{2}{c|}{$\rightarrow$} & \multicolumn{2}{c|}{$\leftarrow$} & \multicolumn{2}{c|}{$\rightarrow$} & \multicolumn{2}{c|}{$\leftarrow$} &  &  \\ \cline{2-17}
 & \textbf{B-4} & \multicolumn{1}{c|}{\textbf{EM}} & \multicolumn{1}{c}{\textbf{B-4}} & \multicolumn{1}{c|}{\textbf{EM}} & \textbf{B-4} & \multicolumn{1}{c|}{\textbf{EM}} & \multicolumn{1}{c}{\textbf{B-4}} & \multicolumn{1}{c|}{\textbf{EM}} & \textbf{B-4} & \multicolumn{1}{c|}{\textbf{EM}} & \multicolumn{1}{c}{\textbf{B-4}} & \multicolumn{1}{c|}{\textbf{EM}} & \textbf{B-4} & \multicolumn{1}{c|}{\textbf{EM}} & \multicolumn{1}{c}{\textbf{B-4}} & \multicolumn{1}{c|}{\textbf{EM}} &  &  \\ \hline
\textbf{Ours} & \multicolumn{1}{r}{71.16} & \multicolumn{1}{r|}{13.2} & 72.70 & 27.2 & \multicolumn{1}{r}{60.98} & \multicolumn{1}{r|}{10.6} & 69.28 & 24.3 & \multicolumn{1}{r}{71.39} & \multicolumn{1}{r|}{15.7} & 72.28 & 26.3 & \multicolumn{1}{r}{74.53} & \multicolumn{1}{r|}{24.3} & 72.43 & 28.5 & \multicolumn{1}{r}{70.59} & \multicolumn{1}{r}{21.26} \\ \hdashline
$\backslash$\textbf{SCLM} & \multicolumn{1}{r}{67.70} & \multicolumn{1}{r|}{11.3} & 68.50 & 23.4 & \multicolumn{1}{r}{55.98} & \multicolumn{1}{r|}{7.5} & 64.39 & 21.6 & \multicolumn{1}{r}{68.05} & \multicolumn{1}{r|}{11.4} & 68.03 & 24.1 & \multicolumn{1}{r}{72.52} & \multicolumn{1}{r|}{20.1} & 68.56 & 24.8 & \multicolumn{1}{r}{66.72} & \multicolumn{1}{r}{18.03} \\
$\backslash$\textbf{PTLM} & \multicolumn{1}{r}{66.91} & \multicolumn{1}{r|}{10.4} & 67.66 & 23.9 & \multicolumn{1}{r}{55.84} & \multicolumn{1}{r|}{7.5} & 63.87 & 21.6 & \multicolumn{1}{r}{67.71} & \multicolumn{1}{r|}{11.3} & 66.86 & 23.5 & \multicolumn{1}{r}{71.91} & \multicolumn{1}{r|}{19.6} & 67.98 & 24.2 & \multicolumn{1}{r}{66.09} & \multicolumn{1}{r}{17.75} \\ \hline
\end{tabular}%
}
\caption{Ablation results on documentation translation.}
\label{tab:t2t-ablation}
     \end{subtable}  \vskip 1mm
     \hfill
    \begin{subtable}[h]{\textwidth}
    \centering
   \resizebox{0.6\textwidth}{!}{%
    \begin{tabular}{l|rr|rr|ll}
    \hline
    \multirow{2}{*}{\textbf{Model}} & \multicolumn{2}{c|}{\textbf{Refine small}} & \multicolumn{2}{c|}{\textbf{Refine medium}} & \multicolumn{2}{c}{\textbf{Avg.}} \\ \cline{2-7} 
    & \textbf{B-4} & \textbf{EM} & \textbf{B-4} & \textbf{EM} & \multicolumn{1}{r}{\textbf{B-4}} & \multicolumn{1}{r}{\textbf{EM}} \\ \hline
    \textbf{Ours} & 80.09 & 13.21 & 91.20 & 2.22 & 85.65 & 7.72 \\ \hdashline
    $\backslash$\textbf{SCLM} & 79.65 & 13.04 & 91.19 & 2.17 & 85.42 & 7.61 \\
    $\backslash$\textbf{PTLM} & 79.73 & 11.31 & 91.13 & 1.68 & 85.43 & 6.50 \\ \hline \hline
    \end{tabular}%
    }
    \caption{Ablation results on code repair.}
    \label{tab:c2c-ablation}
     \end{subtable}
     \caption{Ablation results of downstream tasks, including multilingual code summarization (a), text-to-code generation (b), documentation translation (c), and code repair (d).}
     \label{tab:ablation}
\end{table*}

\subsection{Chinese code summarization}
\label{ap:zh-test}
\paragraph{Data curation} To expand the evaluation on Chinese code summarization, we release a translated variant of mCoNaLa dataset via crowd-sourcing. Specifically, we hire human translators who satisfy all following three criteria to undertake the crowd-sourcing: 
\begin{enumerate}
    \item Must be a native Chinese speaker;
    \item Holding at least a master's degree in Spanish, Japanese, and Russian translation, literature, or related subjects;
    \item Holding professional translation certificates in the corresponding language.
\end{enumerate}

After human translation, we also employ professional engineers who are Chinese native speakers with at least five years of experience in Python to perform further translation refinement. We will release this dataset to speed up the research on multilingual code summarization.

\begin{figure*}[]
     \centering
     \begin{subfigure}[b]{\textwidth}
         \centering
         \includegraphics[width=\textwidth]{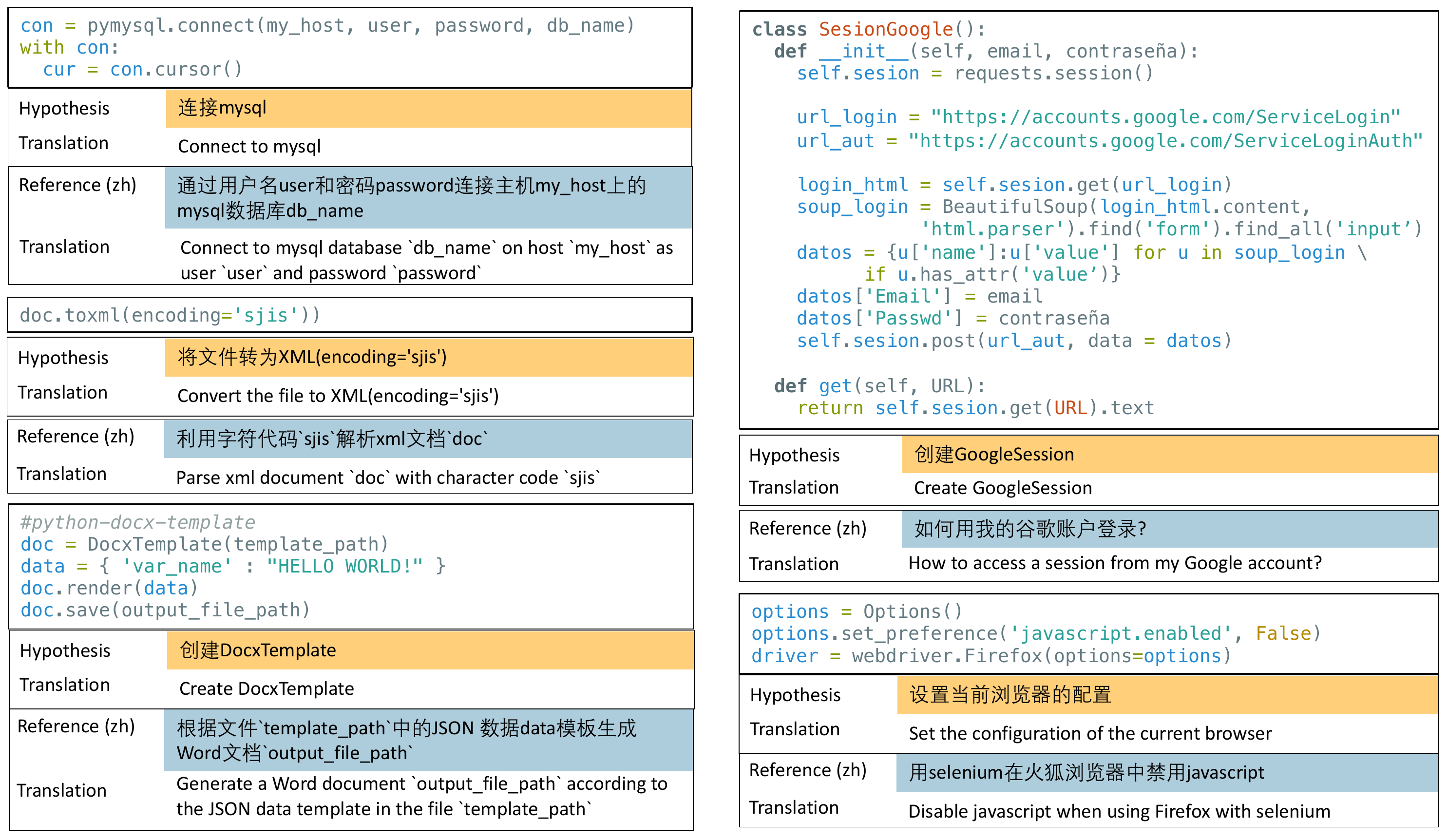}
         \label{fig:c2t-zh-1}
     \end{subfigure}
     \hfill
     \begin{subfigure}[b]{\textwidth}
         \centering
         \includegraphics[width=\textwidth]{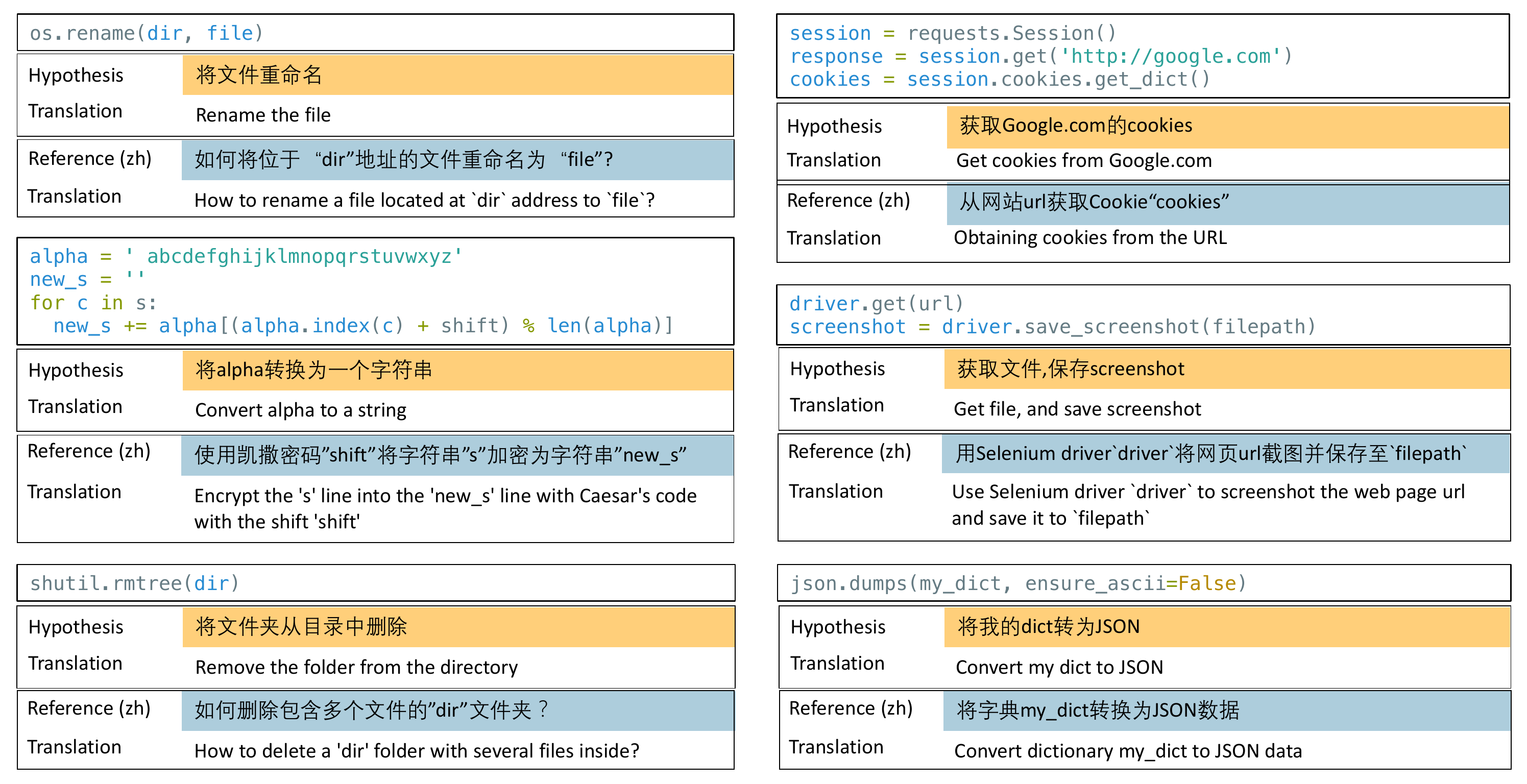} 
         \label{fig:c2t-zh-2}
     \end{subfigure} 
        \caption{Examples of Chinese code summarization with zero-shot prompting.}
        \label{fig:c2t-zh-cases}
\end{figure*}

\paragraph{Examples of Chinese code summarization (zero-shot prompting)} We show the Chinese code summarization examples of our model under zero-shot prompting evaluation in Figure~\ref{fig:c2t-zh-cases}. We prepend the instruction prompt ``translate Python to Chinese: \verb|\n|'' for training and evaluation. It demonstrates that our model equips the zero-shot ability on Chinese code summarization, affirming the positive effect of our cross-lingual pre-training. Moreover, as shown in Figure~\ref{fig:c2t-zh-cases}, our model focuses on the high-level meaning of the input code fragments, neglecting the implementation details. We guess this is because we use code search corpus as NL-PL bilingual training data, where NL instructions comprising high-level descriptions are usually extracted from code comments. It causes a discrepancy between the training and evaluation settings.

\begin{figure*}[]
\begin{center}
\includegraphics[width=\textwidth]{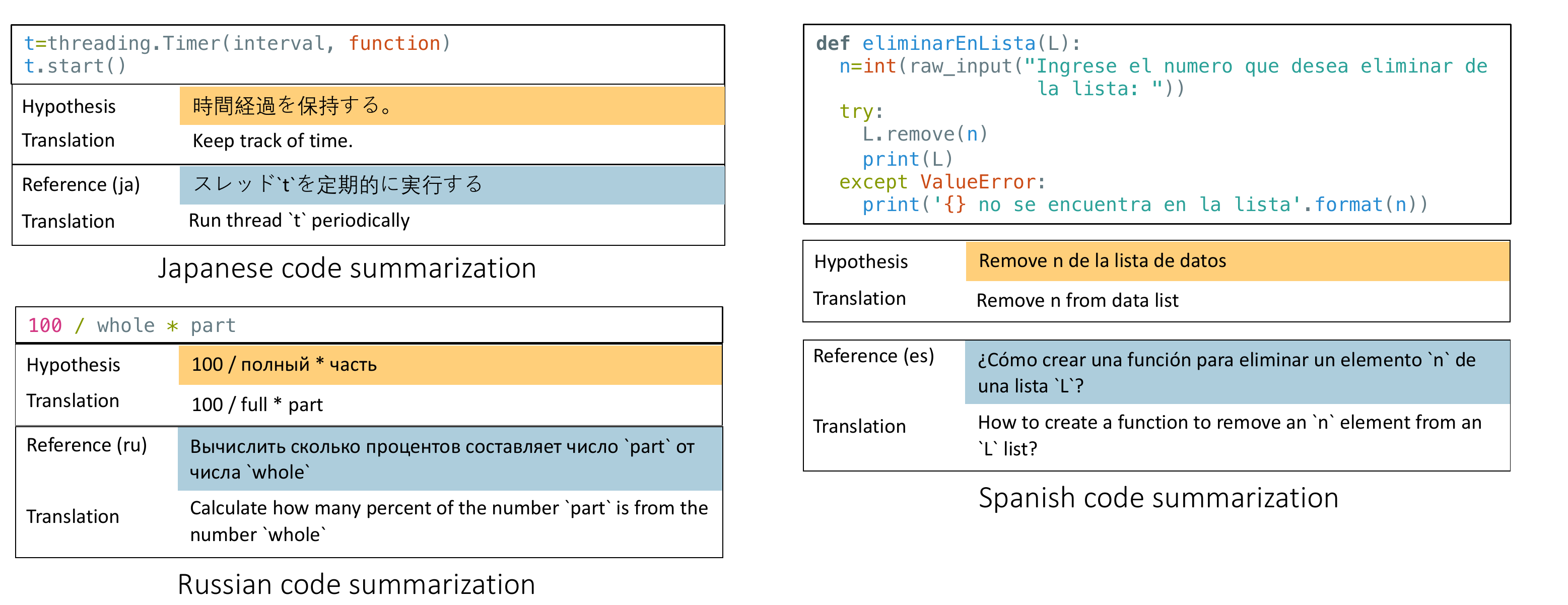}
\caption{Multilingual code summarization (code-to-text) examples with zero-shot prompting. }
\label{fig:c2t-case-2}
\end{center}
\end{figure*}

 \begin{figure}[h]
\begin{center}
\includegraphics[width=\columnwidth]{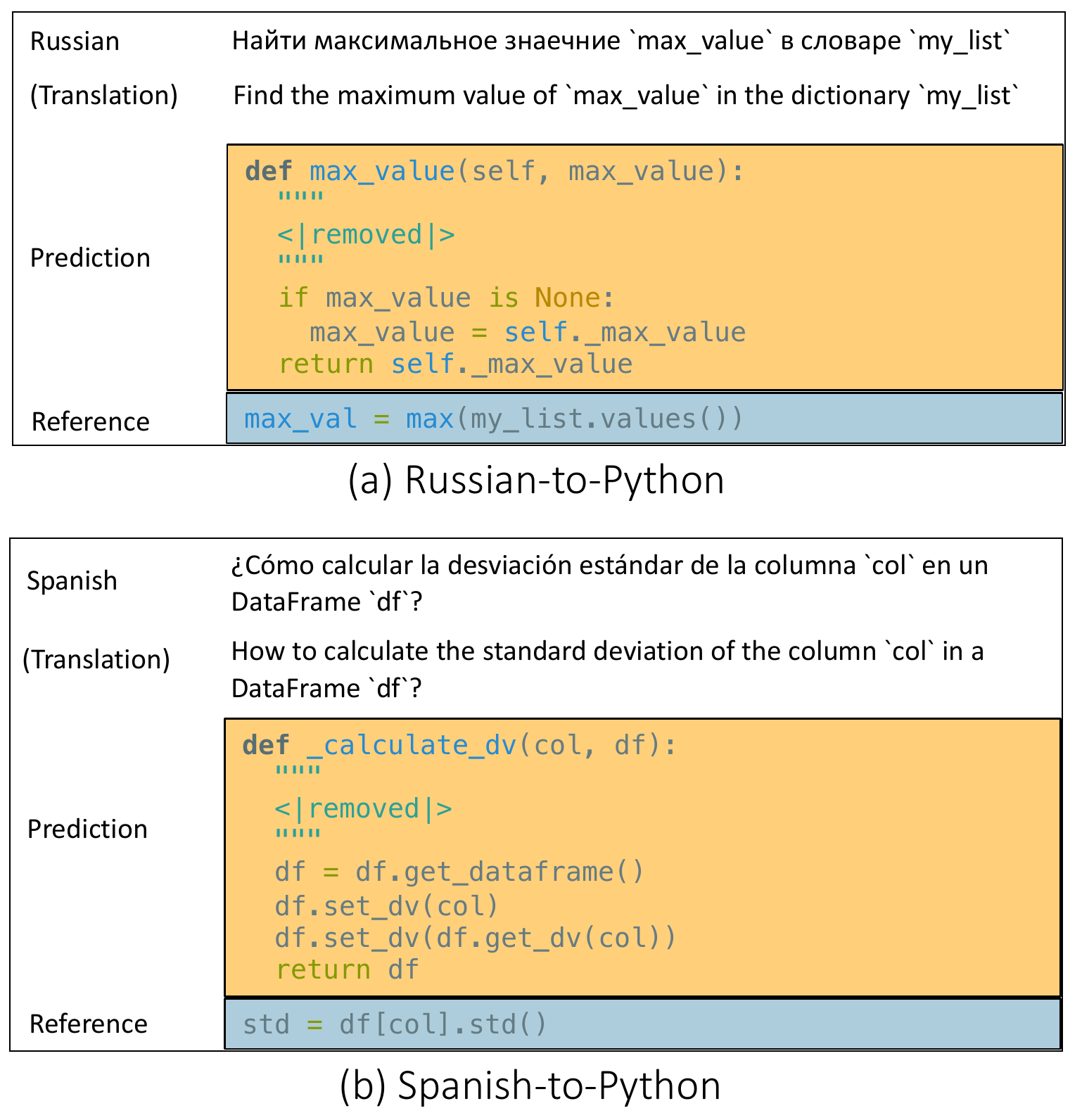}
\caption{Multilingual code summarization (text-to-code) examples with zero-shot prompting. }
\label{fig:t2c-case-2}
\end{center}
\end{figure}

\subsection{Qualitative examples (zero-shot prompting)}
\label{ap:examples}

\begin{figure}[h]
\begin{center}
\includegraphics[width=\columnwidth]{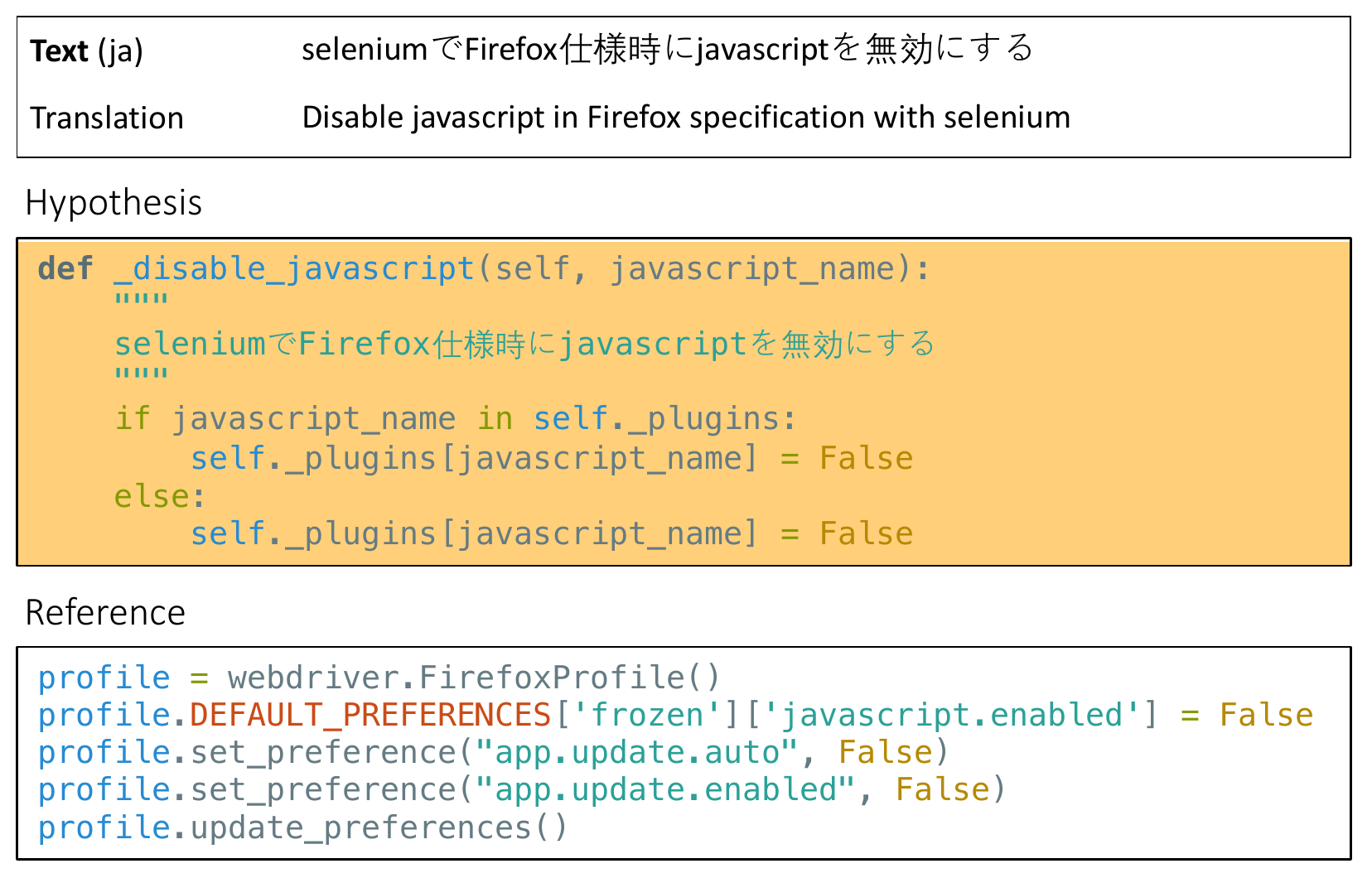}
\vskip -1mm
\caption{Examples of zero-shot multilingual text-to-code generation (Japanese). The region highlighted in orange is a hypothesis generated by our model.}
\label{fig:t2c-case}
\end{center}
\vskip -3mm
\end{figure}

 \paragraph{Zero-shot multilingual PL-to-NL generation} Figure~\ref{fig:c2t-zh-cases} and \ref{fig:c2t-case-2} illustrate the code summarization examples with zero-shot prompting. As mentioned earlier, 
 As illustrated in Figure~\ref{fig:c2t-zh-cases} and \ref{fig:c2t-case-2}, we find that our model focuses on the global overview of code semantics rather than verbalizing the implementation process. Moreover, when explaining a short snippet of code, different people may interpret it with various meanings, which we refer to as ``\textit{program ambiguity}'', making difficulties in annotating and evaluating the multilingual code summarization. This is because the test-set reference of mCoNaLa is human-rewritten, while the training NL is not. 
 We also find that the model tends to copy small code snippets for code summarization. For instance, given inputs ``\# -*- utf-8 -*- '', our model tends to copy the original string rather than describe its usage using NL.

\paragraph{Zero-shot NL-to-PL generation}
Figure~\ref{fig:t2c-case-2} and \ref{fig:t2c-case} demonstrate examples of zero-shot text-to-code generation. 
We also observe that ERNIE-Code is well-performed in generating function names, arguments, and docstrings. It tends to generate function-level snippets and call user-defined functions following the object-oriented logic while lacking the knowledge of builtin functions or user-defined contexts given multilingual NL inputs. 
The given Japanese instruction requires the model to memorize the API usage of \verb|selenium|\footnote{\url{https://selenium-python.readthedocs.io/}} library that our model may never see in the training data. We argue that training on data collected from GitHub and StackOverflow would confer benefits in memorizing and comprehending the API usage and instruction contexts.  
We suspect that the training on additional PL data from GitHub and StackOverflow rather than limited data of CodeSearchNet can lead to large improvements. Note that the generated ``<|removed|>'' docstring in Figure~\ref{fig:t2c-case-2} is consistent with our preprocessing in \S\ref{ap:pl_data}.

\begin{figure}[h]
\begin{center}
\includegraphics[width=\columnwidth]{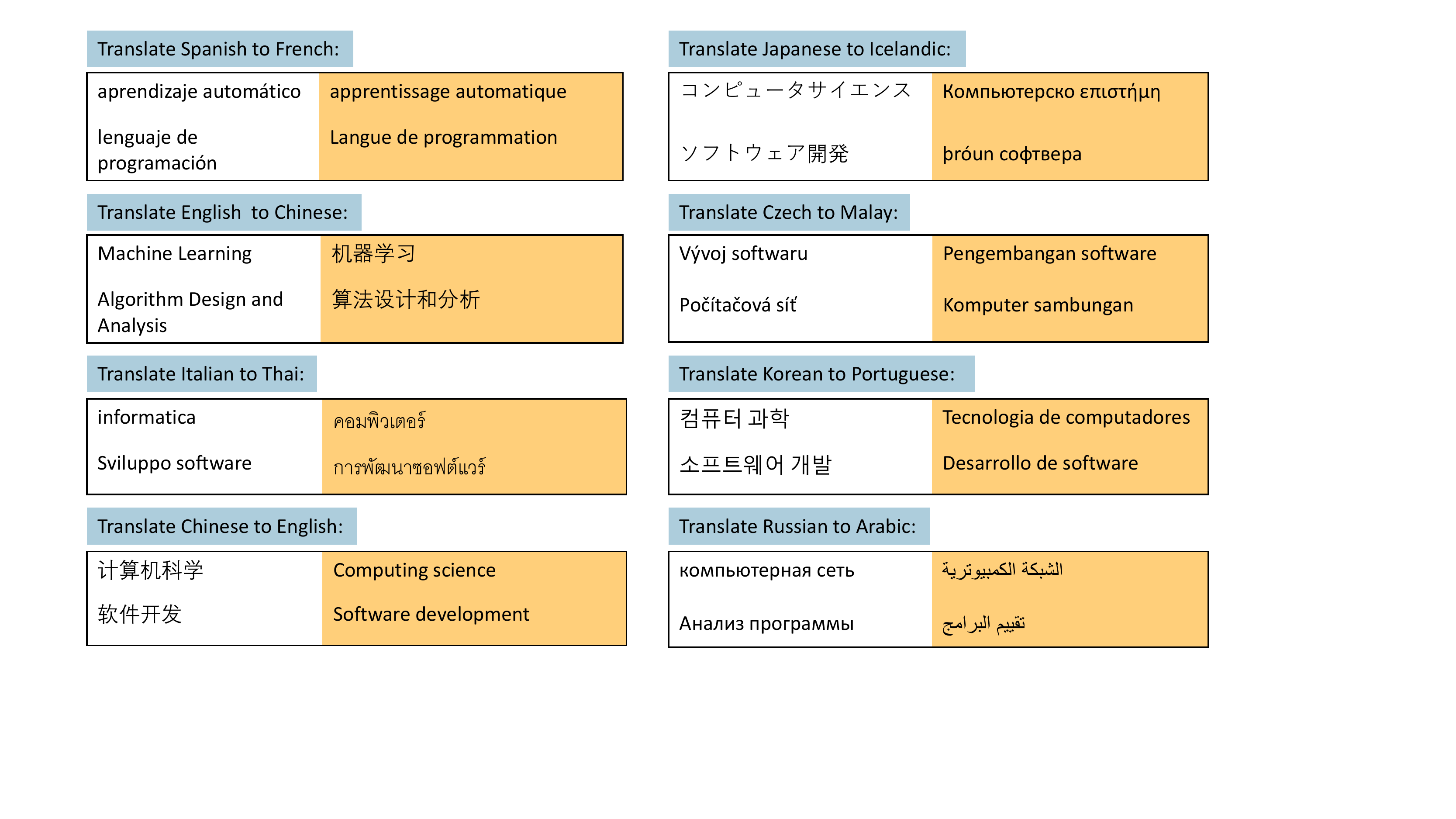}
\caption{Examples of zero-shot text-to-text translation on technical jargon. The region highlighted in orange is the target language, whereas that in blue is the prefixed prompt we use for zero-shot translation.}
\label{fig:mt-case}
\end{center}
\end{figure}

\paragraph{Zero-shot multilingual NL-to-NL translation} To further validate the zero-shot translation capability between multilingual NLs, we report several selected language pairs from different language families and translate technical terminologies with zero-shot prompting. Figure~\ref{fig:mt-case} exhibits examples of multilingual NL translation in eight randomly selected directions, such as Spanish to French and Russian to Arabic. This suggests that our cross-lingual pre-training can capture semantic alignment without seeing direct supervision from bilingual phrase or short-term pairs.

\begin{figure}[]
\vskip -2mm
\begin{center}
\includegraphics[width=\columnwidth]{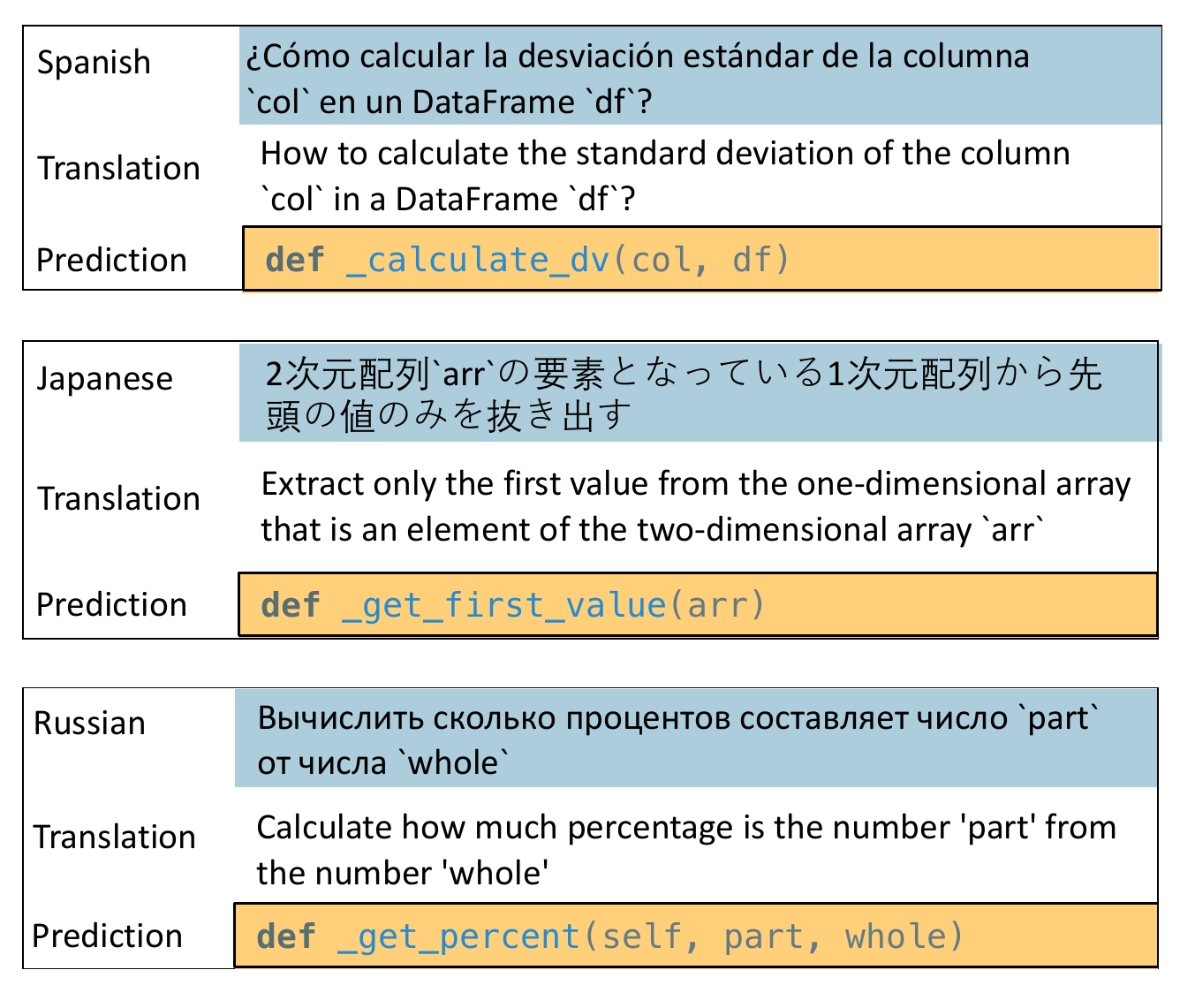}
\vskip -1mm
\caption{Examples of function naming and argument filling in text-to-code generation (zero-shot).}
\label{fig:func_name}
\end{center}
\vskip -2mm
\end{figure}

\paragraph{Qualitative findings} We also observe that our model allows for naming functions and completing corresponding arguments according to multilingual textual instructions, as shown in Figure~\ref{fig:func_name}, confirming that our model learns to bridge the semantics and syntax between multilingual NL instructions and PL functions. 

\end{document}